\documentclass[letterpaper]{article} % DO NOT CHANGE THIS
\usepackage[preprint]{aaai2027}  % DO NOT CHANGE THIS
\usepackage[hyphens]{url}  % DO NOT CHANGE THIS
\usepackage{graphicx} % DO NOT CHANGE THIS
\usepackage{natbib}  % DO NOT CHANGE THIS AND DO NOT ADD ANY OPTIONS TO IT
\usepackage{caption} % DO NOT CHANGE THIS AND DO NOT ADD ANY OPTIONS TO IT
\usepackage{algorithm}
\usepackage{algorithmic}
\usepackage{amsmath,amssymb,amsthm}
\DeclareMathOperator*{\argmin}{arg\,min}
\usepackage{multirow}

\usepackage{newfloat}
\usepackage{listings}
\DeclareCaptionStyle{ruled}{labelfont=normalfont,labelsep=colon,strut=off} % DO NOT CHANGE THIS
\floatstyle{ruled}
\newfloat{listing}{tb}{lst}{}
\floatname{listing}{Listing}

\usepackage{booktabs}

\title{Imagining Recovery: Inference-Time Counterfactual Realignment for Vision-Language-Action Models}

\author{
    Yanyan Zhang,
    Disheng Liu,
    Kai Ye,
    Chaoda Song,
    Xinpeng Li,\\
    Mohsen Hariri,
    Vikash Singh,
    Yu Yin,
    Vipin Chaudhary
}

\affiliations{
    Case Western Reserve University\\
    Cleveland, OH, USA\\
    yxz3106@case.edu
}

\begin{document}

\maketitle

\begin{abstract}

Vision-language-action (VLA) models have improved the flexibility and generality of robotic manipulation, yet they remain fragile to online disruptions, such as changes in task goal, scene configuration, or robot state. Existing recovery methods often require failure data, policy retraining, or external corrective agents, introducing additional data requirements and execution risks. We propose Counterfactual Realignment (CoRe), a training-free framework that recovers a frozen VLA at inference time without failure data. Upon detecting a deviation, CoRe imagines how the policy would continue toward the current goal from a recent viable state, using synthesized observations in place of physical execution, and then minimally realigns the robot and scene to rejoin this imagined continuation before returning control to the policy. Recovery is therefore planned without physical trial-and-error, preserves completed task progress, and handles both mid-episode instruction changes and physical perturbations in a unified manner. Extensive experiments across multiple simulators, VLA backbones, and real-world settings show that CoRe improves success rates by up to 85.0 percentage points to near-nominal levels while reducing physical restorations by 42.2\%, without policy fine-tuning or failure-specific recovery training.

\end{abstract}


\section{Introduction}

By coupling large-scale vision-language representations with robot action generation, vision-language-action (VLA) models offer a promising path toward general-purpose robots that follow diverse instructions across objects, tasks, and environments~\cite{rt2,openvla,Ma_2026}. Yet, trained almost exclusively on successful, forward-progressing demonstrations~\cite{RACER,duan2025aha}, VLAs remain brittle to out-of-distribution (OOD) execution states~\cite{pumacay2024colosseum,robustvla}. Once the joint consistency of language, observation, and robot configuration breaks, the policy lacks corrective behavior to recover autonomously~\cite{Fei_2026_CVPR,xiao2026selfimproving}, and closed-loop execution compounds minor errors into task failure~\cite{HYDRA,liu2025what}. Such deviations are routine in deployment, where physical disturbances alter the scene or robot state and instructions change mid-task~\cite{liu2025what,Hi_Robot}.

\begin{figure}[!tb]
    \centering
    \includegraphics[width=0.9\linewidth]{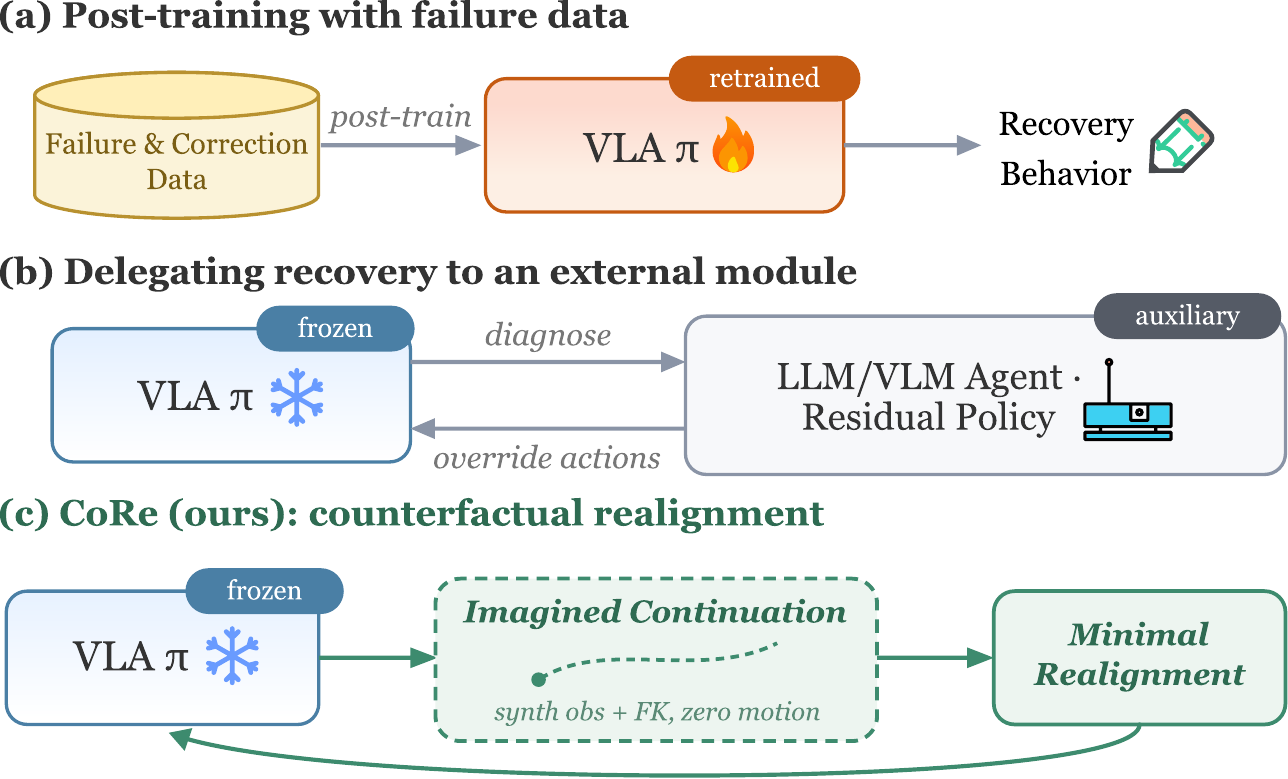}
    \caption{\textbf{OOD recovery paradigms for VLA models.} (a) Post-training on failure data and (b) delegating recovery to an external module both hinge on costly retraining or a reliable auxiliary system. (c) CoRe instead plans recovery virtually and returns control to the same frozen policy.}
    \label{fig:compare}
\end{figure}

\begin{figure*}[!t]
    \centering
    \includegraphics[width=0.9\textwidth]{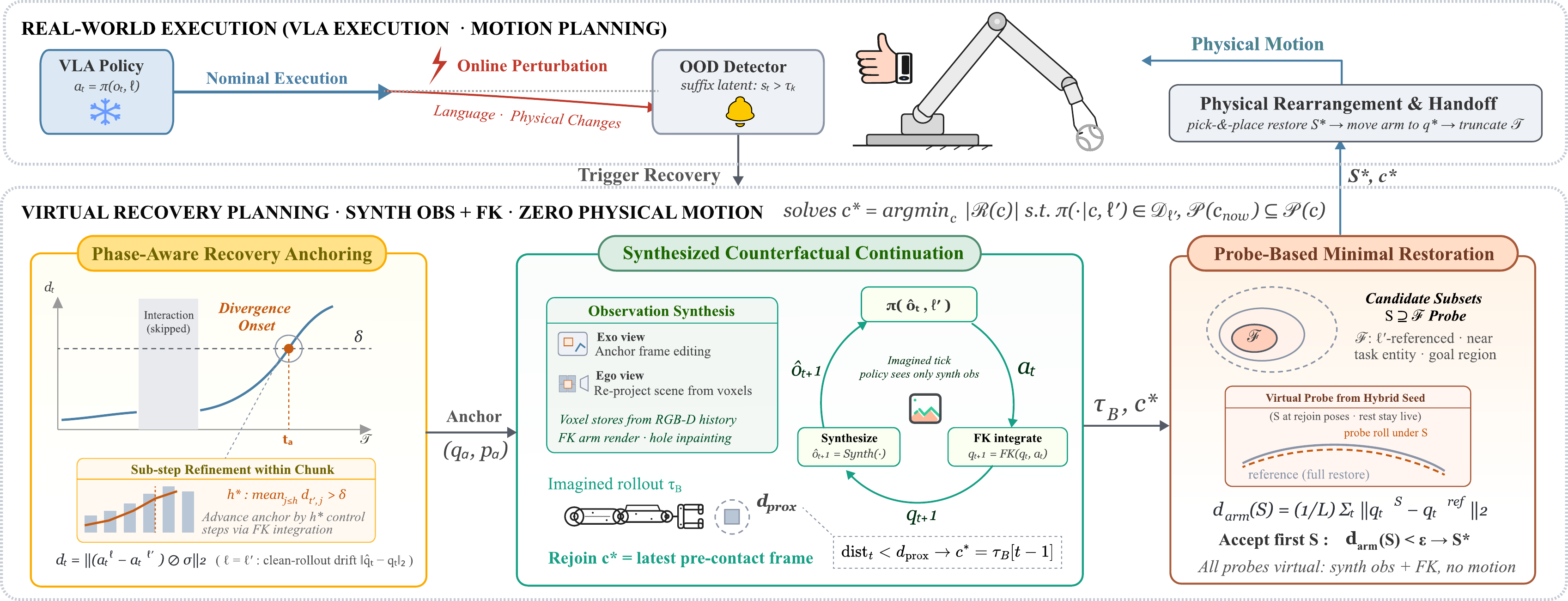}
    \caption{\textbf{Overview of CoRe.} An online perturbation trips the suffix OOD detector and interrupts the frozen VLA (top). Recovery runs through the virtual planning stages (bottom), passing the recovery anchor, the rejoin state $c^*$, and the restoration subset $S^*$ left to right. Only the final rearrangement and handoff produces physical motion before control returns to the frozen VLA.}
    \label{fig:pipeline}
\end{figure*}

Recognizing this limitation, recent work explores failure correction and recovery for VLAs along two main directions (Fig.~\ref{fig:compare}). The first augments nominal demonstrations with failure-conditioned supervision, collected from human interventions during deployment~\cite{wu2026flowpro}, from perturbations injected into successful trajectories with corrective segments appended~\cite{Zhao_2026_CVPR}, or from failed rollouts synthesized through simulators, learned world models, and policy-driven exploration~\cite{lin2025failsafe, li2026learning, xiao2026selfimproving}, and then post-trains the policy through behavioral cloning~\cite{Zhao_2026_CVPR, xiao2026selfimproving}, preference optimization~\cite{wu2026flowpro}, or related objectives~\cite{liufu2026repo}. Such methods require costly failure-data collection and may require repeated collection and retraining as new failure modes emerge~\cite{wu2026flowpro,berk2026recall}. The second direction delegates recovery to an auxiliary module, including LLM/VLM-based agents that diagnose failures and replan~\cite{liu2023reflect}, external supervisors that issue textual, visual, or executable corrections~\cite{ye2025robofac,lin2025failsafe, zeng2025diagnose}, and learned residual policies that modify actions at runtime~\cite{hu2026recovla}. Such modules add training or inference overhead and must themselves produce reliable, executable recovery decisions~\cite{lin2025failsafe,ye2025robofac,hu2026recovla}. Beyond these two lines, resetting or rewinding execution to a predetermined state, whether a fixed subtask checkpoint or a frame retrieved from prior demonstrations~\cite{ma2026cyclevla, shin2026back, zhao2026retrieve}, can discard completed progress, fail to escape the OOD state, or remain confined to states already in the repertoire.

We propose \textbf{Counterfactual Realignment (CoRe)}, an inference-time framework that recovers a frozen VLA from online language switches, environmental disturbances, and robot state interference without retraining, failure-specific data, or auxiliary models, as shown in Fig.~\ref{fig:pipeline}. The central idea is a \emph{counterfactual continuation}, a short closed-loop rollout of what the VLA \emph{would} do under the current target from a recent in-distribution state, from which a minimal physical recovery plan is derived. CoRe detects OOD states from the VLA's own action-expert latent, anchors recovery at the latest in-distribution frame that preserves completed progress, and rolls out the policy from there using observations synthesized from accumulated RGB-D history and forward kinematics, requiring neither ground-truth access nor physical robot motion. \textbf{The rollout yields a safe rejoin state,} and virtual probes over progressively smaller restoration subsets select the minimum set of displaced objects whose restoration preserves the imagined trajectory. Physical motion occurs only after this validated plan is established, replacing the physical trial-and-error of iterative correction methods with virtual planning. Our key contributions are:

\begin{itemize}
\item We propose CoRe, a training-free framework that recovers frozen VLAs from online perturbations through counterfactual continuation and minimal physical realignment.

\item We show that VLA policies are robust to geometrically synthesized observations, so closed-loop counterfactual rollouts suffice for recovery planning without a learned world model or privileged environment access.

\item Extensive experiments across simulators, VLA backbones, and real-world settings show consistent success gains under diverse online perturbations without policy fine-tuning.

\end{itemize}

\section{Related Work}

\paragraph{Vision-Language-Action Models.} VLA models adapt pre-trained vision-language backbones to action prediction~\cite{rt2, openvla, liu2024rdt}. Autoregressive designs co-tokenize discretized actions with language~\cite{rt2, openvla}, while continuous-action architectures generate high-frequency action chunks through flow matching or diffusion for dexterous and bimanual control~\cite{pi0, liu2024rdt}. As these models lack built-in recovery, CoRe keeps the policy frozen and builds recovery externally without modifying its architecture or training.

\paragraph{Failure Detection and Recovery for VLA Models.} One line augments training data with failure-conditioned supervision, injecting perturbations into successful trajectories and training corrective behaviors~\cite{Zhao_2026_CVPR, lin2025failsafe, li2026learning} or fine-tuning with subtask progress labels to enable proactive backtracking~\cite{ma2026cyclevla}. The other delegates recovery to an external module, an LLM/VLM agent that diagnoses failures and replans~\cite{liu2023reflect, ye2025robofac}, a learned residual policy~\cite{hu2026recovla}, or an uncertainty-aware detector that triggers replanning~\cite{xu2025can, gu2026safe}. As discussed above, the former requires costly failure collection and retraining~\cite{Zhao_2026_CVPR, li2026learning} while the latter stands or falls with the auxiliary system itself~\cite{gu2026safe}. Recent work further shows that VLA internal features separate successful from failed rollouts, enabling lightweight failure detection. CoRe shares this insight but goes beyond detection, constructing the full pipeline from anchor identification through counterfactual planning to minimal physical intervention without retraining the policy, learning a separate recovery model, or relying on an external reasoning agent.

\begin{figure}[!tb]
    \centering
    \includegraphics[width=0.73\linewidth]{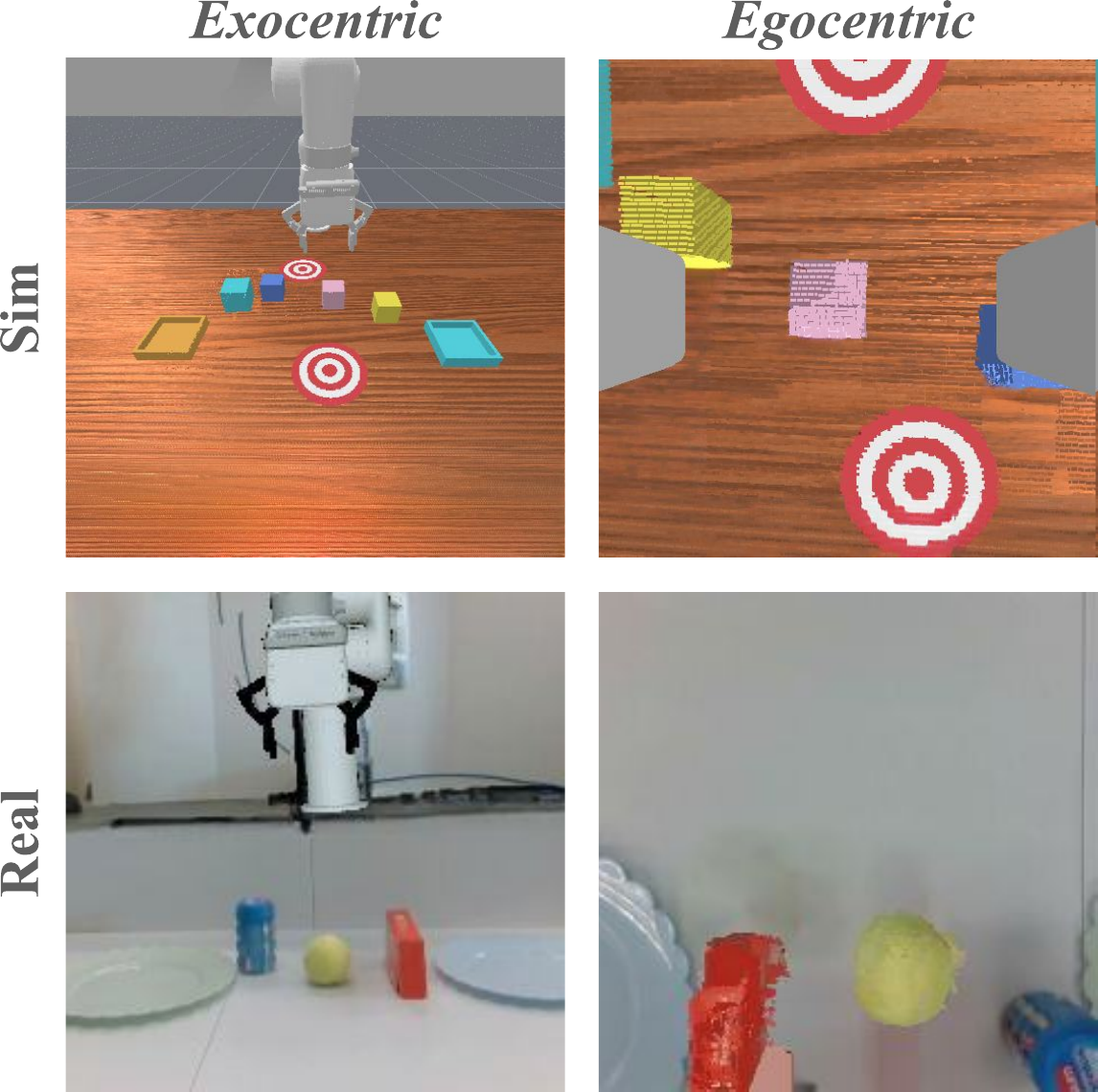}
    \caption{\textbf{Synthesized observations.} Exocentric and egocentric views at imagined ticks, arm rendered by FK, residual holes inpainted.}
    \label{fig:synth_example}
\end{figure}

\section{Methodology}

A VLA policy $\pi$ maps an observation $o_t$, RGB images from one or more cameras plus a language instruction $\ell$, to an action chunk $a_t$ of $N$ actions executed open-loop before the next query. During execution we record a \emph{trace} $\mathcal{T} = \{(o_t, a_t, q_t, p_t)\}_{t=0}^{T}$ of observations, actions, arm joint configurations $q_t$, and scene object poses $p_t = \{(x_i, r_i)\}$. An \emph{online perturbation} at time $t_p$ (\textit{e.g.,} a mid-execution instruction change $\ell \to \ell'$ or a physical disturbance altering $p_t$ or $q_t$), drives execution off the policy's distribution $\mathcal{D}$ while $\pi$ stays frozen. Recovery seeks a target configuration $c^* = (q^*, p^*)$, an arm pose with a possibly partial scene arrangement, that minimizes physical effort
\begin{equation}
\begin{gathered}
c^* = \argmin_{c \in \mathcal{C}} \; |\mathcal{R}(c)| \\
\text{s.t.}\quad
\pi(\cdot \mid c,\ell') \in \mathcal{D}_{\ell'},\quad
\mathcal{P}(c_{\text{now}}) \subseteq \mathcal{P}(c).
\end{gathered}
\label{eq:objective}
\end{equation}
where $|\mathcal{R}(c)|$ counts the object manipulations required to reach $c$, the first constraint keeps the VLA's continuation under $\ell'$ on the in-distribution manifold $\mathcal{D}_{\ell'}$, and the second, with $\mathcal{P}(c)$ the subgoals completed at $c$, preserves all accumulated task progress. CoRe solves Eq.~\ref{eq:objective} entirely at inference time, as illustrated in Fig.~\ref{fig:pipeline}.

\subsection{Online Suffix OOD Detection}
\label{sec:detection}

Following SAFE~\cite{gu2026safe}, we monitor the \emph{action-expert suffix latent}, the last hidden state of the action head before motor decoding. At each policy query, $K$ stochastic forward passes produce $K$ latent vectors that are mean-pooled into $h_t \in \mathbb{R}^{d}$. We then compute a per-task Mahalanobis score
\begin{equation}
s_t = \sqrt{(z_t - \bar{z}_k)^\top \widetilde{\Sigma}_k^{-1} (z_t - \bar{z}_k)}
\label{eq:mahala}
\end{equation}
where $z_t = (h_t - \mu_k) \oslash \sigma_k$ is the z-standardized feature and $\oslash$ denotes element-wise division. Here $\mu_k$ and $\sigma_k$ are the per-dimension mean and standard deviation from in-distribution rollouts for task $k$, $\bar{z}_k$ and $\widetilde{\Sigma}_k$ are the mean and ridge-regularized covariance of the standardized features, and $\tau_k$ is a high quantile of $s_t$ calibrated offline on held-out successes. A trigger fires when $s_t > \tau_k$ for $n_{\text{consec}}$ consecutive queries. % [FIX 5] "control steps" -> "queries"
% Unlike SAFE~\cite{gu2026safe}, which trains a neural classifier and aborts on detection, CoRe uses a calibration-based threshold and treats detection as the entry point for autonomous recovery.
Unlike SAFE~\cite{gu2026safe}, which trains a neural classifier and aborts execution upon detection, CoRe calibrates a threshold solely from successful rollouts and uses each trigger to initiate autonomous realignment. This design avoids training a separate detector and goes beyond failure containment: it preserves completed progress, realigns the system, and resumes task execution autonomously.

\subsection{Phase-Aware Recovery Anchoring}
\label{sec:anchor}

Upon detection, CoRe identifies a \emph{recovery anchor}, the latest trace frame from which recovery can begin while preserving completed progress. Scanning $\mathcal{T}$ forward from the most recent recovery endpoint, or from $t=0$ on the first recovery, it compares the chunks predicted at each frame under the two instructions, one forward pass each,
\begin{equation}
d_{t,j} = \big\| \big(a_{t,j}^{\ell} - a_{t,j}^{\ell'}\big) \oslash \sigma \big\|_2,
\label{eq:divstep}
\end{equation}
where $a_{t,j}^{\ell}$ is the $j$-th action predicted under $\ell$ and $\sigma$ holds per-dimension in-distribution action standard deviations. The frame-level divergence is then
\begin{equation}
d_t =
\begin{cases}
d_{t,0} & \text{if } \ell \neq \ell' \\[2pt]
\big\| q_t^{\text{clean}} - q_t^{\text{trace}} \big\|_2 & \text{if } \ell = \ell'
\end{cases}
\label{eq:div}
\end{equation}
using the first, immediately executed action. When the instruction is unchanged the two chunks coincide, so $d_t$ instead compares a clean rollout against the recorded configuration. The scan is \emph{phase-aware}, skipping frames until displaced objects or articulated joints settle and requiring gripper clearance from every movable object, so the anchor lies in reproducible free space. The first frame with $d_t > \delta$ marks the divergence onset, with a fallback to the most recent in-distribution frame whose short rollout under $\ell'$ stays clean. As the policy is queried per chunk, the onset is localized only to chunk resolution, so CoRe refines it within the chunk executed from the last sub-threshold frame $t'$, whose per-step divergences are already available, cutting at the first control step whose running mean crosses $\delta$
\begin{equation}
h^{*} = \min\Big\{\, h \;:\; \tfrac{1}{h+1}\textstyle\sum_{j=0}^{h} d_{t',j} \,>\, \delta \,\Big\},
\label{eq:substep}
\end{equation}
and advancing the anchor arm $h^{*}$ control steps from $t'$ by forward-kinematic integration of that chunk, inheriting its static pre-interaction scene.

\begin{figure*}[!t]
    \centering
    \includegraphics[width=0.85\textwidth]{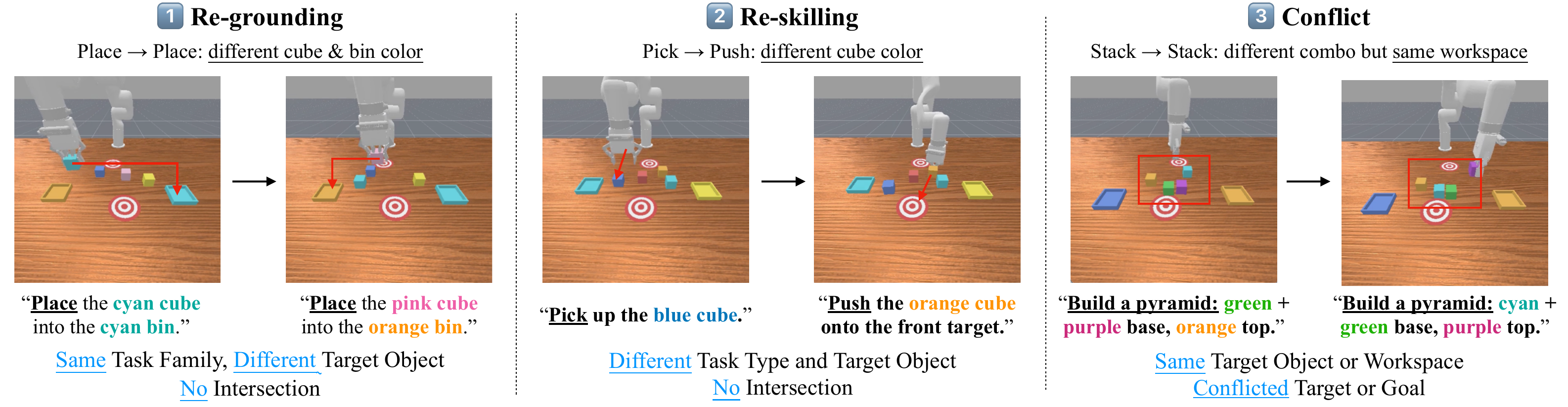}
    \caption{\textbf{Overview of the LangSwitch Tasks}, spanning the \textit{Re-grounding}, \textit{Re-skilling}, and \textit{Conflict} switch families defined in the Experimental Setup.}
    \label{fig:langshift_tasks}
\end{figure*}

\begin{table*}[!t]
  \centering
  \resizebox{\textwidth}{!}{%
  \begin{tabular}{c l c ccc ccc ccc cc}
  \toprule
  \multirow{2}{*}{}
  & \multirow{2}{*}{\textbf{Method}}
  & \multirow{2}{*}{\textbf{No-Switch}}
  & \multicolumn{3}{c}{\textbf{Re-grounding}}
  & \multicolumn{3}{c}{\textbf{Re-skilling}}
  & \multicolumn{3}{c}{\textbf{Conflict}}
  & \multicolumn{2}{c}{\textbf{Average}} \\
  \cmidrule(lr){4-6} \cmidrule(lr){7-9} \cmidrule(lr){10-12} \cmidrule(lr){13-14}
  & &
  & Pk$\to$Pk & Ps$\to$Ps & Pc$\to$Pc
  & Pk$\to$Ps & Pc$\to$Pl & St$\to$Pk
  & Ps$\to$Pl & Pc$\to$Ps & St$\to$St
  & Switch & All \\
  \midrule
  \multirow{3}{*}{\rotatebox[origin=c]{90}{\textit{Base}}}
  & $\pi_0$~\citep{pi0}                       & 90.6 & 8.3$\pm$13.6 & 0.7$\pm$0.6 & 0.0$\pm$0.0 & 7.7$\pm$9.0 & 12.3$\pm$7.5 & 27.0$\pm$20.9 & 0.7$\pm$1.2 & 91.0$\pm$8.2 & 17.0$\pm$12.1 & 18.3$\pm$5.9 & 44.1 \\
  & $\pi_{0.5}$~\citep{Intelligence202505AV}  & \textbf{97.8} & 29.0$\pm$50.2 & 21.3$\pm$37.0 & 0.0$\pm$0.0 & 33.0$\pm$39.8 & 56.3$\pm$5.5 & 65.7$\pm$56.9 & 7.0$\pm$12.1 & 85.0$\pm$11.5 & 22.3$\pm$17.1 & 35.5$\pm$21.3 & 57.8 \\
  & GR00T N1.7~\citep{youliang2026NVIDIA}                   & 88.4 & 5.0$\pm$2.6 & 4.0$\pm$5.3 & 0.0$\pm$0.0 & 4.7$\pm$5.7 & 27.7$\pm$1.2 & 12.7$\pm$11.0 & 7.7$\pm$12.4 & 88.7$\pm$4.0 & 17.0$\pm$15.6 & 18.6$\pm$4.6 & 43.5 \\
  \midrule
  \multirow{3}{*}{\rotatebox[origin=c]{90}{\textit{Ours}}}
   & $\pi_0$ + CoRe        & 90.0 & \textbf{98.7$\pm$1.5} & \underline{87.7$\pm$1.5} & 73.0$\pm$2.6 & \underline{95.7$\pm$0.6} & 88.7$\pm$2.9 & \textbf{95.3$\pm$4.7} & \underline{83.3$\pm$8.1} & \underline{97.0$\pm$2.0} & \underline{68.0$\pm$7.9} & \underline{87.5$\pm$0.6} & \underline{88.4} \\
  & $\pi_{0.5}$ + CoRe    & \underline{97.2} & \underline{97.7$\pm$0.6} & \textbf{98.3$\pm$1.5} & \textbf{87.0$\pm$1.0} & \textbf{99.7$\pm$0.6} & \textbf{96.0$\pm$1.0} & \underline{92.7$\pm$1.5} & \textbf{92.3$\pm$7.4} & \textbf{98.7$\pm$1.5} & \textbf{72.0$\pm$11.8} & \textbf{92.7$\pm$1.3} & \textbf{94.3} \\
  & GR00T N1.7 + CoRe     & 90.2 & 93.7$\pm$3.5 & 76.0$\pm$4.6 & \underline{74.7$\pm$2.9} & 89.7$\pm$5.0 & \underline{89.3$\pm$2.9} & 87.7$\pm$9.3 & \underline{83.3$\pm$1.5} & 91.7$\pm$2.5 & 57.7$\pm$5.7 & 82.6$\pm$2.1 & 85.3 \\
  \bottomrule
  \end{tabular}%
  }
    \caption{\textbf{Success rate (\%) on the LangSwitch tasks with an xArm6 robot.} Each switch pair is evaluated at 3 switch timings (shallow, mid, deep), 100 episodes each; per-pair cells and the Switch average report the mean $\pm$ sample standard deviation across the 3 timings. No-Switch reports the mean over the 5 base skills (100 episodes each), and All is the macro average over all 14 task settings (5 No-Switch $+$ 9 switch pairs). Pk=Pick, Ps=Push, Pc=Place, Pl=Pull, St=Stack.}
  \label{tab:langswitch}
\end{table*}

\subsection{Synthesized Counterfactual Continuation}
\label{sec:counterfactual}

From the recovery anchor, CoRe rolls out the frozen VLA $\pi$ under the target instruction $\ell'$ to construct a counterfactual continuation $\tau_B$, a short trajectory of what the policy \emph{would} do from an in-distribution state, built entirely on \textbf{synthesized observations} rather than ground-truth renders or physical robot motion, so no simulator access is required.

\paragraph{Observation synthesis.}  Each imagined tick must show the VLA the imagined state, with the arm at the pose its imagined actions would have reached, namely the FK-computed pose, and objects possibly away from their real positions. CoRe maintains per-object voxel stores in each object's local frame, accumulated from segmented RGB-D observations, plus a world-frame background store. The \textbf{third-person (exocentric) view} from the fixed camera is synthesized by editing the real frame at the anchor. Pixels of the arm and displaced objects are removed, the vacated regions are filled by projecting background voxels, displaced objects are re-rendered at their imagined poses through z-buffered voxel splatting, and the arm is composited from a forward-kinematic render. The \textbf{first-person (egocentric) view} moves with the imagined arm pose, so the entire image is instead reconstructed from the voxel stores at the imagined viewpoint. Residual holes in both views are filled by neural or classical inpainting (Fig.~\ref{fig:synth_example}).

\paragraph{Closed-loop rollout.} Each imagined tick synthesizes observations for both cameras, queries $\pi$ with them under $\ell'$, and advances the arm by forward kinematics
\begin{equation}
q_{t+1} = \textsc{FK}(q_t, a_t), \quad \hat{o}_{t+1} = \textsc{Synth}(q_{t+1}, p_t)
\label{eq:fk}
\end{equation}
where $\textsc{FK}$ applies the robot-specific forward-kinematic integration of action $a_t$ at configuration $q_t$ and $\hat{o}_{t+1}$ is the synthesized observation at the updated configuration. The rollout terminates once the tool-center-point comes within a proximity threshold $d_{\text{prox}}$ of any scene object, beyond which contact dynamics cannot be modeled by FK alone. The latest pre-contact frame becomes the \textbf{rejoin state} $c^*$ to which the real system will be realigned, keeping the rejoin state in free space while preserving as much of the generated approach as possible. Backbones that execute long action chunks fully open loop must additionally keep a larger clearance, since an entire chunk runs without re-observation after handoff.

\subsection{Probe-Based Minimal Restoration}
\label{sec:subset}

Not every displaced object needs physical restoration, since an object far from the task-relevant workspace may not affect the continuation from the rejoin state. CoRe therefore selects the \textbf{minimum subset} whose restoration is necessary. An object counts as \emph{displaced} when its pose deviates from the rejoin state $c^*$ beyond a position or orientation tolerance ($\epsilon_{\text{pos}}$, $\epsilon_{\text{rot}}$). A \emph{forced set} $\mathcal{F}$ is always restored, covering objects referenced by $\ell'$, objects within a radius $r_{\text{veto}}$ of any task-relevant entity, and objects occupying a designated goal region, while the remaining displaced objects are candidates for exclusion. Each candidate subset $S \supseteq \mathcal{F}$ is evaluated by a \emph{virtual probe}, a counterfactual rollout from a hybrid seed in which objects in $S$ sit at their rejoin poses and all others keep their current live poses. The probe trajectory is compared to a full-restore reference through
\begin{equation}
d_{\text{arm}}(S) = \frac{1}{L} \sum_{t=1}^{L} \| q_t^{S} - q_t^{\text{ref}} \|_2
\label{eq:armdiv}
\end{equation}
where $q_t^{S}$ and $q_t^{\text{ref}}$ are the arm configurations under $S$ and the reference and $L = \min(|\tau_S|, |\tau_{\text{ref}}|)$ counts the comparable frames. Probing subsets in order of increasing size from $\mathcal{F}$ alone, CoRe accepts the first with $d_{\text{arm}}(S) < \epsilon$, a trajectory-equivalence threshold.

\begin{figure*}[!t]
    \centering
    \includegraphics[width=0.9\textwidth]{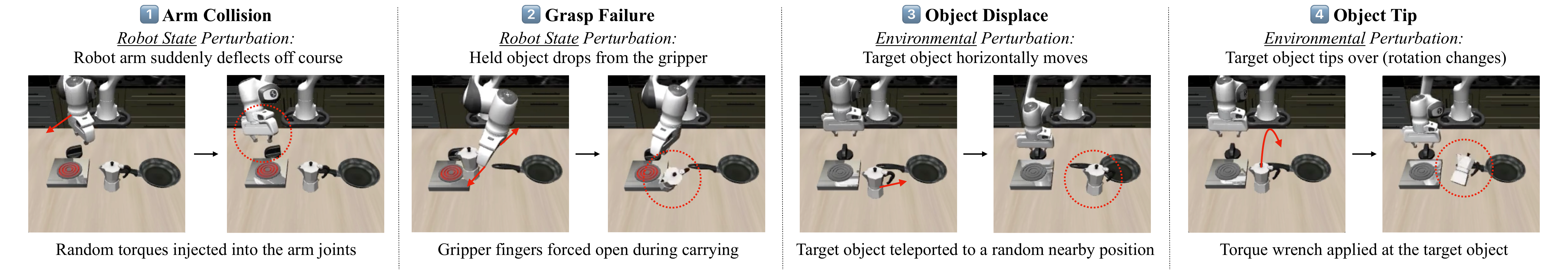}
    \caption{\textbf{Overview of the LIBERO-Long with Perturbation Tasks}, spanning the \textit{Arm Collision}, \textit{Grasp Failure}, \textit{Object Displace}, and \textit{Object Tip} perturbation families defined in the Experimental Setup.}
    \label{fig:libero_preturb}
\end{figure*}

\begin{table*}[!t]
  \centering
  \small
  \begin{tabular*}{\textwidth}{@{\extracolsep{\fill}} c l cccc c @{}}
  \toprule
  & \multirow{2}{*}{\textbf{Method}}
  & \multicolumn{4}{c}{\textbf{Perturbation Type}}
  & \multirow{2}{*}{\textbf{Average}} \\
  \cmidrule(lr){3-6}
  & & Arm Coll. & Grasp Fail. & Obj. Disp. & Obj. Tip & \\
  \midrule
  \multirow{6}{*}{\rotatebox[origin=c]{90}{\textit{Baselines}}}
  & OpenVLA~\citep{openvla}                  & 3.3$\pm$2.5 & 41.3$\pm$1.6 & 3.2$\pm$1.0 & 3.3$\pm$0.6 & 12.8$\pm$0.3 \\
  & OpenVLA-OFT~\cite{kim2025fine}           & 18.7$\pm$4.3 & 65.0$\pm$9.5 & 13.5$\pm$6.4 & 10.3$\pm$3.9 & 26.9$\pm$1.8 \\
  & $\pi_{0.5}$~\citep{Intelligence202505AV} & 37.0$\pm$8.7 & 65.0$\pm$1.7 & 43.8$\pm$7.0 & 28.0$\pm$2.8 & 43.5$\pm$2.3 \\
  & VLA-Adapter~\citep{adapter}              & 21.3$\pm$5.8 & 65.0$\pm$4.5 & 10.8$\pm$1.9 & 10.5$\pm$2.8 & 26.9$\pm$2.4 \\
  & X-VLA~\citep{zheng2026xvla}              & 60.5$\pm$5.2 & 51.2$\pm$7.3 & 6.2$\pm$0.6 & 9.7$\pm$0.8 & 31.9$\pm$2.8 \\
  & GR00T N1.7~\citep{youliang2026NVIDIA}    & 15.7$\pm$6.3 & 46.0$\pm$4.8 & 16.3$\pm$3.5 & 15.5$\pm$2.6 & 23.4$\pm$2.5 \\
  \cmidrule(lr){2-7}
  & RobustVLA~\citep{robustvla}              & 13.2$\pm$5.1 & 47.2$\pm$11.9 & 2.7$\pm$1.0 & 3.3$\pm$0.8 & 16.6$\pm$4.3 \\
  \midrule
  \multirow{3}{*}{\rotatebox[origin=c]{90}{\textit{Ours}}}
  & $\pi_{0.5}$ + CoRe    & \textbf{89.0$\pm$2.8} & \textbf{90.2$\pm$3.0} & \underline{87.5$\pm$1.0} & \textbf{86.3$\pm$2.1} & \textbf{88.2$\pm$0.8} (81.9) \\
  & X-VLA + CoRe          & \underline{84.8$\pm$3.8} & \underline{81.7$\pm$1.3} & \textbf{88.3$\pm$4.3} & \underline{83.0$\pm$5.2} & \underline{84.5$\pm$2.6} (79.0) \\
  & GR00T N1.7 + CoRe     & 74.3$\pm$8.5 & 71.5$\pm$0.5 & 72.2$\pm$4.4 & 61.2$\pm$13.3 & 69.8$\pm$3.3 (65.0) \\
  \bottomrule
  \end{tabular*}
    \caption{\textbf{Post-perturbation success rate (\%) on LIBERO-Long with a Franka Panda robot.} Each of the 3 perturbation timings (shallow, mid, deep) is evaluated over 200 episodes (10 tasks $\times$ 20 episodes); cells report mean $\pm$ sample standard deviation across the 3 timings. For CoRe, the Average additionally reports in parentheses the fully autonomous \emph{Arm-Only} variant, which weights each restoration by the arm's measured execution feasibility.}
  \label{tab:libero}
\end{table*}

\subsection{Physical Rearrangement and Handoff}
\label{sec:recovery}

Only this final stage moves the robot. Each object in $S^*$ is returned to its rejoin pose through motion-planned pick-and-place in top-first height order, every placement is verified to within $\epsilon_{\text{verify}}$ with one retry, and unrecoverable cases such as a failed grasp or an unreachable pose are reported as a structured intervention request. The arm is then moved to the rejoin configuration $q^*$, with backbone-specific handoff steps deferred to the Appendix, and control returns to the frozen $\pi$. The execution trace is truncated to the recovery anchor, so any later recovery builds on this one's progress rather than regressing to earlier states.

\section{Experiments}

\begin{table}[!tb]
\centering
\small
\begin{tabular*}{\columnwidth}{
  @{\extracolsep{\fill}}lccc@{}
}
\toprule
Condition & $\pi_{0.5}$ & +CoRe & $\Delta$ [n (pp)] \\
\midrule
\multicolumn{4}{@{}l}{\textit{$\bullet$ Language Switch}} \\
\quad Re-grounding (pick$\to$pick)
  & 6/20 & 18/20 & $+12$ ($+60$) \\
\quad Re-skilling (pick$\to$push)
  & 1/20 & 18/20 & $+17$ ($+85$) \\
\quad Conflict (place$\to$push)
  & 1/20 & 18/20 & $+17$ ($+85$) \\
\midrule
\multicolumn{4}{@{}l}{\textit{$\bullet$ Physical Perturbation}} \\
\quad Arm Collision (place)
  & 9/20 & 17/20 & $+8$ ($+40$) \\
\quad Scene Rearrangement (pick)
  & 4/20 & 19/20 & $+15$ ($+75$) \\
\midrule
\multicolumn{4}{@{}l}{%
  \textit{Nominal}\hfill
  pick 19/20 \qquad place 19/20 \qquad push 20/20} \\
\bottomrule
\end{tabular*}
\caption{\textbf{Real-world xArm6 results}, evaluated using layout-paired trials with stratified perturbation timings.}
\label{tab:real_results}
\end{table}

\begin{figure}[!tb]
    \centering
    \includegraphics[width=0.8\linewidth]{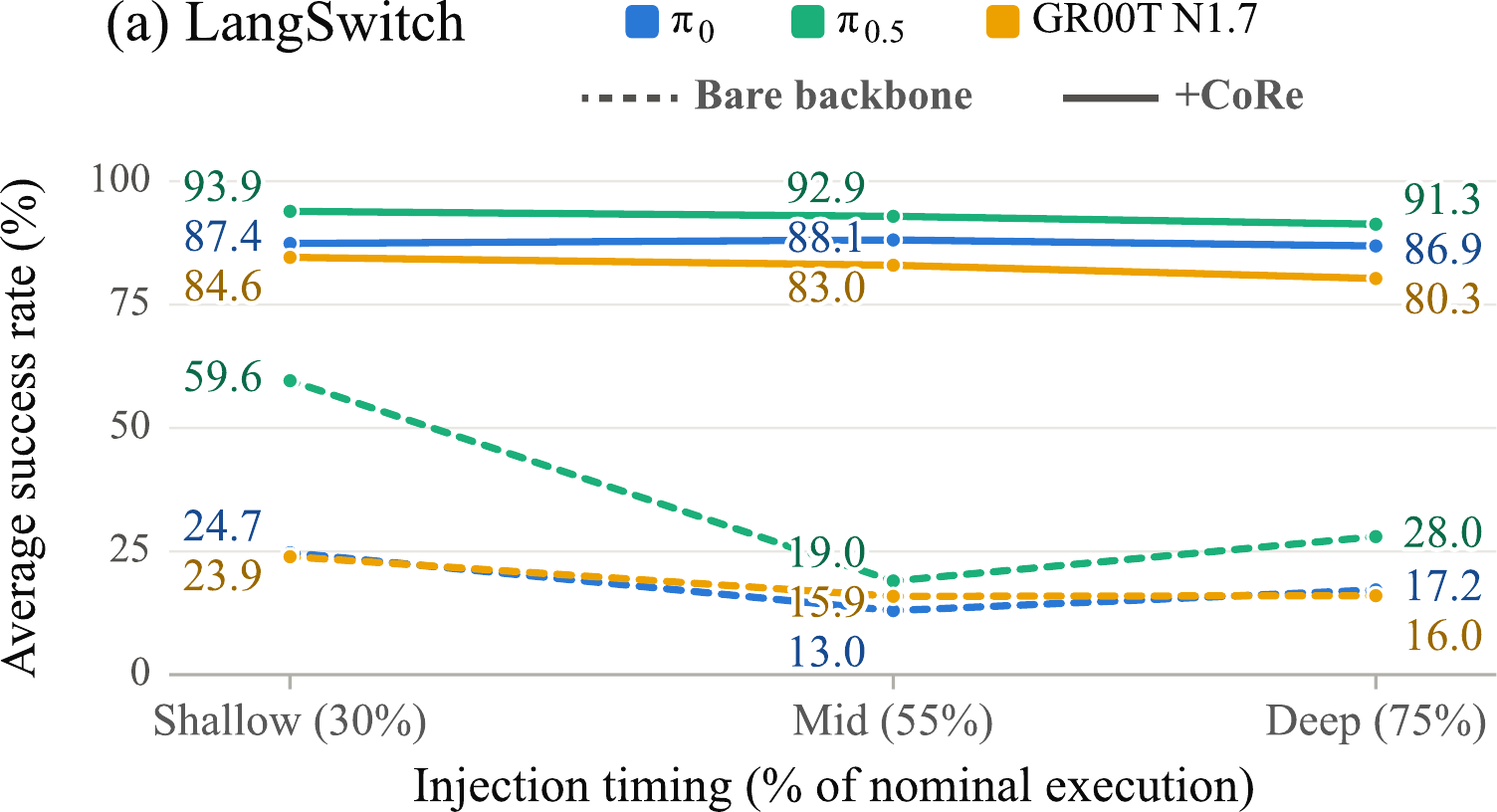}\\[4pt]
    \includegraphics[width=0.8\linewidth]{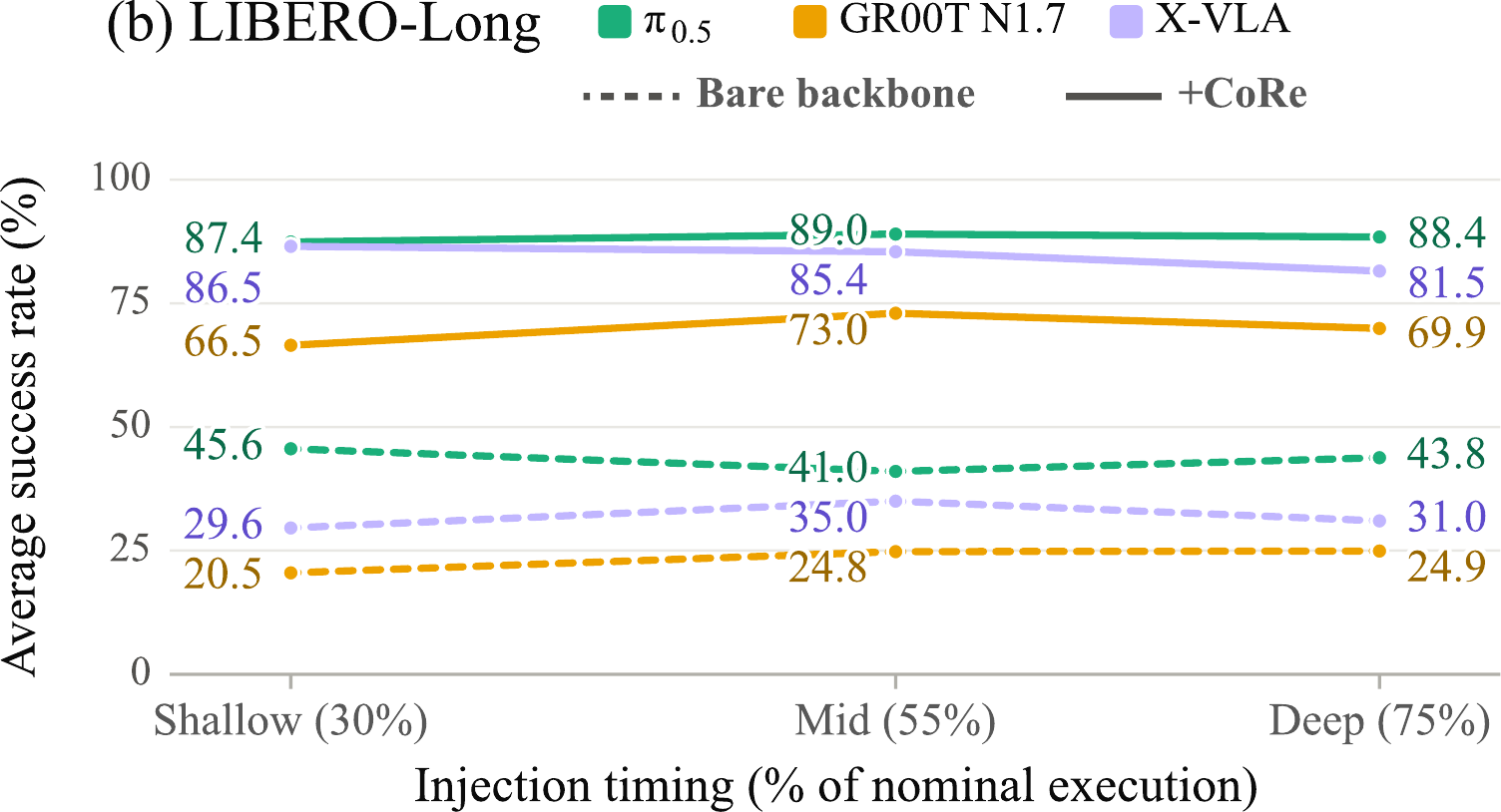}
    \caption{\textbf{Effect of injection timing.} Success rate at each injection timing, bare versus with CoRe. Bare policies are timing-sensitive under language switches yet nearly timing-invariant under physical perturbations, and CoRe lifts and flattens all curves.}
    \label{fig:timing}
\end{figure}

\subsection{Experimental Setup}

\textbf{Simulation.} We evaluate CoRe on two benchmarks that separately target goal switches and physical perturbations across distinct simulators and embodiments. Both vary a perturbation along two axes, its \emph{type} and its \emph{timing}, the latter injected at a shallow, mid, or deep point. \textbf{(1) LangSwitch} (Fig.~\ref{fig:langshift_tasks}) studies \emph{online language switches} with an xArm6 in ManiSkill. Five tasks, placing, pushing, pulling, pick-and-place, and pyramid stacking, share a unified environment, baselines are trained jointly on their demonstrations, and mid-episode the instruction changes from $\ell$ to a new target $\ell'$. Switch types form three families. \emph{Re-grounding} redirects the current skill to a different object, \emph{Re-skilling} replaces the skill with a different one, and \emph{Conflict} issues a goal contradicting the workspace or target state the ongoing task has produced. Timing is taken within the ongoing skill. \textbf{(2) LIBERO-Long with Perturbation} (Fig.~\ref{fig:libero_preturb}) studies \emph{online physical perturbations} with a Franka Panda in robosuite, on the most challenging LIBERO suite of 10 long-horizon tasks. A single perturbation is injected per episode, in four families. \emph{Arm Collision} knocks the arm off course with a random joint torque impulse, \emph{Grasp Failure} forces the gripper open mid-transport and drops the held object, \emph{Object Displace} relocates the target before the arm reaches it, and \emph{Object Tip} topples the target in place with a torque wrench. Timing is taken within the phase each family targets, namely the full episode for collisions, the pre-contact approach for displacement and tipping, and the post-lift carry for grasp failures, with phase boundaries calibrated per backbone from its own clean rollouts so that policies of different speeds are perturbed at comparable stages. Restoration differs between the benchmarks. LangSwitch executes every restoration physically through motion-planned pick-and-place, demonstrating that recovery plans are executable, whereas LIBERO applies restorations by direct state writes and accounts for executability through the grasp-feasibility weighting in Tab.~\ref{tab:libero}, decoupling the quality of recovery decisions from the mechanics of executing them. Injection parameters and timing specifications are given in the Appendix. 

\begin{figure}[!tb]
    \centering
    \includegraphics[width=0.69\linewidth]{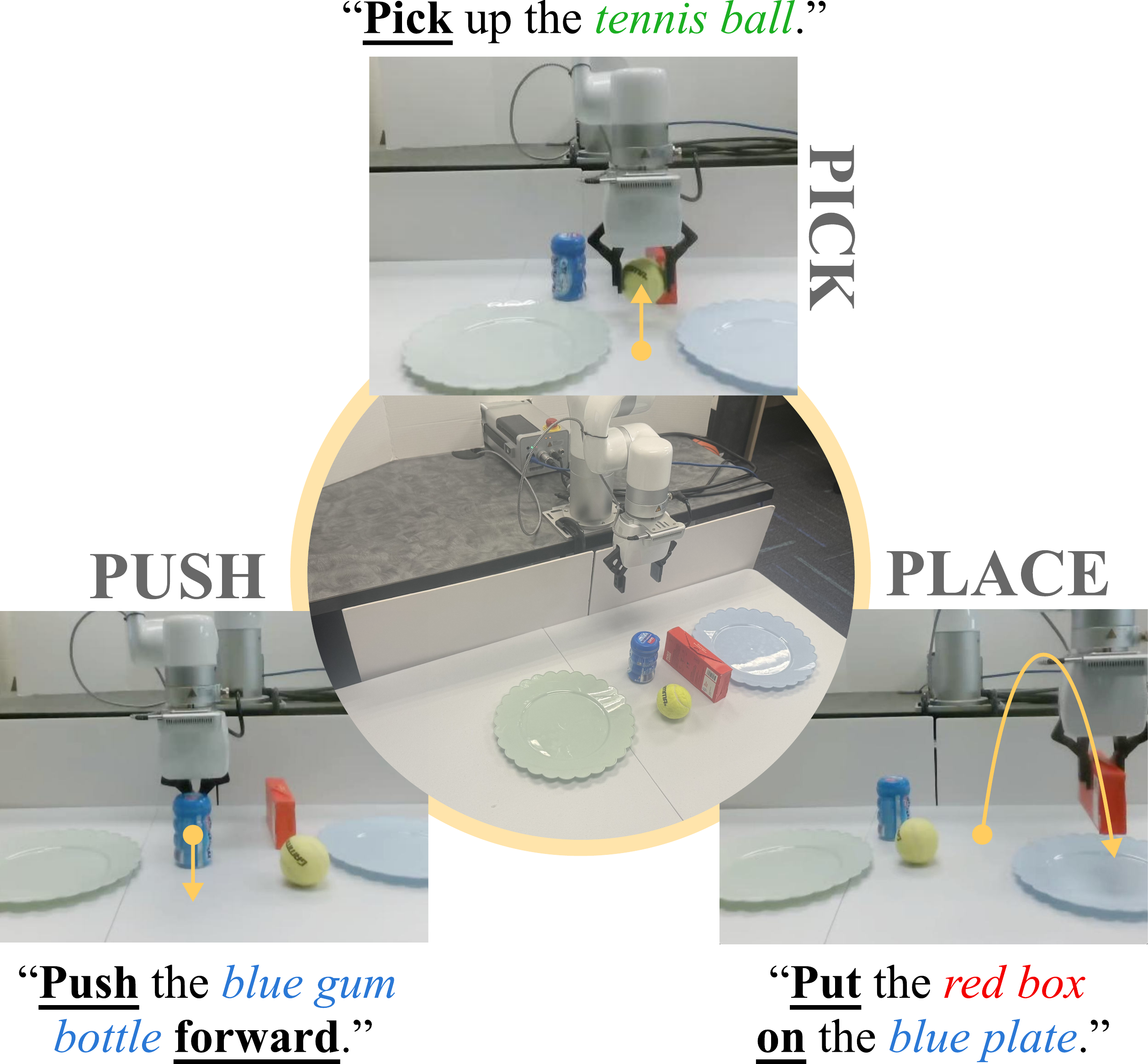}
    \caption{\textbf{Overview of the Real-World Tasks.} We validate the feasibility of CoRe beyond simulation on a physical xArm6 across multiple task families and objects.}
    \label{fig:real_overview}
\end{figure}

\noindent\textbf{Real-World.} We deploy CoRe on a physical xArm6 with two RealSense D435i cameras, one fixed third-person and one wrist-mounted (Fig.~\ref{fig:real_overview}). A $\pi_{0.5}$ backbone is fine-tuned with 100 teleoperated demonstrations per skill, covering pick, place, and push over three household objects and two plates. Five conditions comprise the three LangSwitch families defined above and two physical perturbations, with \emph{Arm Collision} applying a randomized impulse to all six joints and \emph{Scene Rearrangement} having an operator relocate every object midway through the episode. Each condition runs 20 episodes across shallow, mid, and deep timings, with phase boundaries recalibrated on real nominal rollouts. Trials are paired by layout, instruction, and replayed collision impulse.

\subsection{Experimental Results and Analysis}

\textbf{Simulation.} On LangSwitch, CoRe raises the three backbones from 18.3--35.5\% to 82.6--92.7\% average success and improves all 27 switch settings (Tab.~\ref{tab:langswitch}). Conflict remains the hardest family because the new goal contradicts the state produced by the original task. No false triggers occur in 1{,}500 unswitched runs. On LIBERO-Long (Tab.~\ref{tab:libero}), CoRe reaches 69.8–88.2\%, compared with 12.8–43.5\% for bare baselines; the Arm-Only variant retains 65.0–81.9\%. Bare policies are highly timing-sensitive under language switches but nearly timing-invariant under physical perturbations (Fig.~\ref{fig:timing}). Mid-skill language switches often lock the policy onto the object already being manipulated, whereas physical perturbations invalidate the demonstrated approach regardless of prior progress. By handing control back at a validated counterfactual rejoin state, CoRe reduces the timing spread to at most 6.5 points.

\noindent\textbf{Real-World.} As shown in Tab.~\ref{tab:real_results}, the bare $\pi_{0.5}$ completes at most 9/20 episodes under any perturbation despite 19--20/20 nominal competence, collapsing to 1/20 when the switch demands a new skill or conflicts with already-produced scene state. CoRe lifts every condition to 17--19/20 (85--95\%), including 1/20 $\to$ 18/20 on both Re-skilling and Conflict, and the gains hold across all three injection depths. Every recovery here is executed entirely on hardware. Restorations are motion-planned pick-and-place actions on the real scene, and the median recovery realigns only the arm, occupying 8.8\,s of physical motion. This demonstrates CoRe end to end, without ground-truth state or privileged information.

\subsection{Ablation Studies}

\paragraph{Fidelity of Synthesized Counterfactual Rollouts.}
Because recovery decisions rely on synthesized rollouts, we replay them from identical anchors under ground-truth renders and measure endpoint TCP divergence under a reduced protocol detailed in the Appendix. Across all 6 settings in Fig.~\ref{fig:synth_vs_gt}, the median ranges from 0.36 to 3.28\,cm, remaining below the 2\,cm (LangSwitch) and 8\,cm (LIBERO) pre-contact margins used for recovery decisions. A second ground-truth replay controls for the policy's own sampling variability, and on LangSwitch the synthesized rollout stays within 1.2--1.7$\times$ of this floor. The heaviest tail (GR00T on LIBERO, p90 of 9.7\,cm) arises where the two ground-truth replays themselves diverge by 11--41\,cm, so it reflects policy multimodality, and even long-chunk X-VLA stays at one fifth of its 15\,cm rejoin standoff.

\begin{figure}[!tb]
    \centering
    \includegraphics[width=0.75\linewidth]{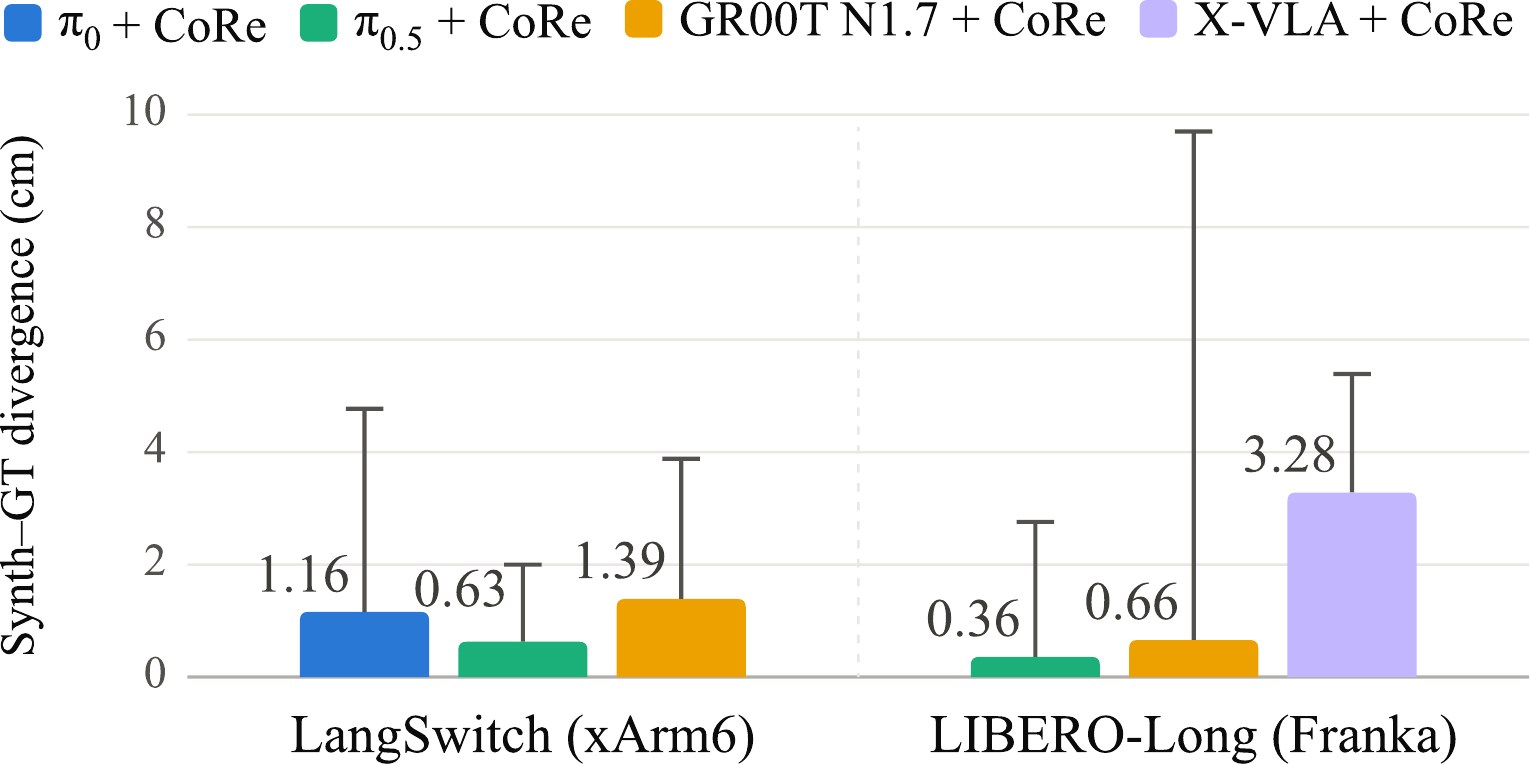}
    \caption{\textbf{Fidelity of synthesized rollouts.} Endpoint TCP divergence between synthesized and GT rollouts from the same anchor (bars: median, whiskers: 90th percentile).}
    \label{fig:synth_vs_gt}
\end{figure}

\begin{figure}[!tb]
  \centering
  \small
  \begin{tabular}{lccc}
    \toprule
    \multirow{2}{*}{\textbf{Method}} & \multirow{2}{*}{\textbf{SR (\%)} $\uparrow$} & \multirow{2}{*}{\textbf{\#Restored} $\downarrow$} & \textbf{Post-hand} \\
     & & & \textbf{Steps} $\downarrow$ \\
    \midrule
    Episode Reset        & 88.5 & 1.18 & 108.7 \\
    Anchor Rollback      & 89.6 & 1.14 & 108.1 \\
    CoRe w/o Sub         & 88.1 & 1.16 & \textbf{74.1} \\
    \textbf{CoRe (ours)} & \textbf{90.7} & \textbf{0.67} & 74.4 \\
    \bottomrule
  \end{tabular}\\[7pt]
  \includegraphics[width=0.8\linewidth]{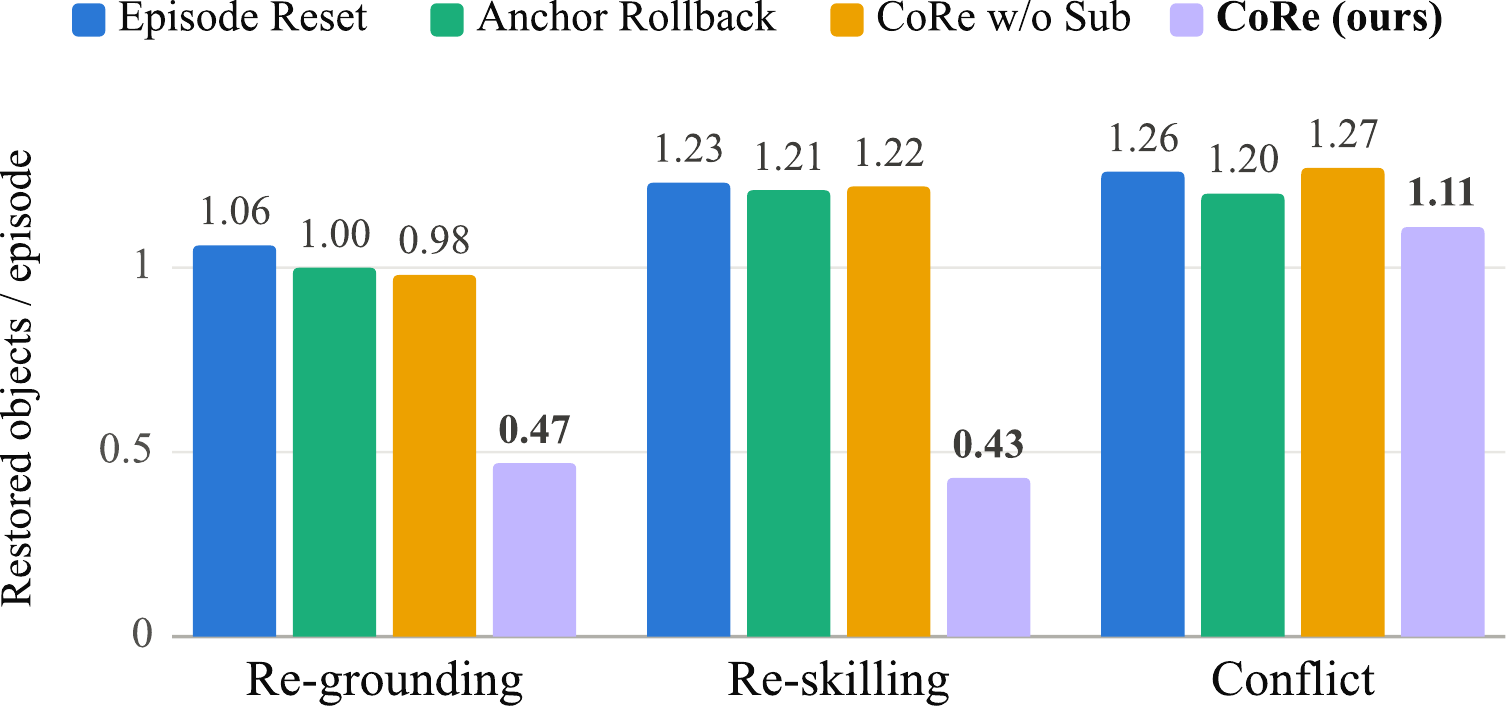}
  \caption{\textbf{Mechanism ablation on LangSwitch.} \emph{Top}, overall comparison of the 4 recovery variants. \#Restored is the number of objects physically moved per episode, and Post-hand Steps counts the control steps executed after handoff. \emph{Bottom}, \#Restored split by switch family.}
  \label{fig:mech_ablation}
\end{figure}

\paragraph{Recovery Mechanism Ablation.}
As shown in Fig.~\ref{fig:mech_ablation}, we compare CoRe with three variants that share the same detection, physical rearrangement, and handoff components but differ in the state to which the system is realigned. \textit{Episode Reset} returns the scene and arm to their initial states, \textit{Anchor Rollback} restores everything at the recovery anchor, and \textit{CoRe w/o Sub} follows the counterfactual continuation but restores every displaced object. All variants share an identical 270-episode LangSwitch protocol on $\pi_{0.5}$ (Appendix), and restoration pruning is additionally measured on the real Scene-Rearrangement suite. \textbf{(a) Necessity of the Counterfactual Continuation.} Counterfactual rejoining reduces post-handoff execution from about 108 to 74 steps while maintaining slightly higher success than Episode Reset and Anchor Rollback, as it hands control back at a rollout-validated pre-contact state rather than an earlier recorded state.  \textbf{(b) Minimal Restoration in Simulation and on Hardware.} CoRe restores 0.67 objects per episode versus 1.16 under full restoration, with comparable post-handoff execution and slightly higher success. It prunes 52--65\% of restorations under Re-grounding and Re-skilling but only 13\% under Conflict, where most displaced objects affect the continuation. On the real suite, where the operator displaces all three objects every episode, CoRe restores 1.15 of 2.77 above-tolerance displacements per recovery and leaves 58\% untouched, squarely within the simulated 52--65\% band. In 11 of 13 recovery episodes it moves exactly the instruction-referenced target, and in the other two the solo-target probe itself rejects the smaller subset with $d_{\text{arm}} = 0.19 > \epsilon$, so the extra object it admits is probe-certified rather than conservatively included.

\begin{figure}[!tb]
    \centering
    \includegraphics[width=\linewidth]{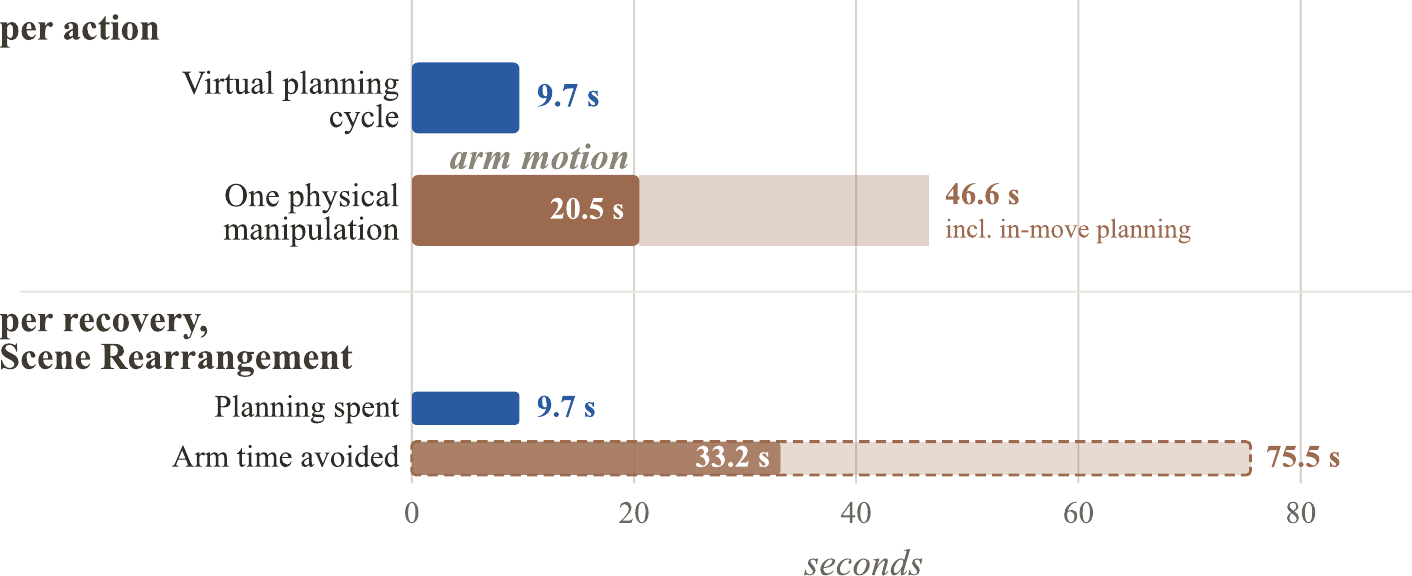}
   \caption{\textbf{Virtual probing versus physical execution.} Probing a recovery hypothesis in imagination is 2-5 times cheaper than testing it with the arm.}
    \label{fig:time_efficiency}
\end{figure}

\paragraph{Cost and Safety of Virtual Probing.}
Virtual probing buys two things over physical trial, time and safety, both measured over the 165 recoveries of the 100 layout-paired real episodes. A complete virtual planning cycle takes a median of 9.7\,s, which amortizes to 1.2\,s per imagined action chunk over the median 8 chunk counterfactual. A single physical manipulation, picking one object up and placing it elsewhere with the arm handoff excluded, occupies the arm for a median of 20.5\,s of motion and 46.6\,s of wall clock including in-move planning and perception. Probing a hypothesis in imagination is thus two to five times cheaper than testing it with the arm, and on Scene Rearrangement the 1.62 manipulations pruned per recovery convert 9.7\,s of planning into 33 to 76\,s of avoided arm time, as summarized in Fig.~\ref{fig:time_efficiency}. The same substitution buys safety. Under identical layouts and replayed injections, 21 of the 100 baseline episodes end in a collision or an operator emergency stop, 27 percent of all baseline failures, whereas CoRe ends with 5, one of them the injected impulse itself striking clutter. Exploratory contact is displaced into imagination, where rollouts truncate at a proximity boundary rather than at the object.

\section{Conclusion}

We presented Counterfactual Realignment (CoRe), a training-free framework that recovers frozen VLAs from online language switches and physical perturbations. By imagining a continuation from a viable prior state, CoRe identifies a pre-contact rejoin state and restores only the objects needed to resume execution, replacing physical trial-and-error with virtual recovery planning. Extensive experiments demonstrate that CoRe restores near-nominal success with fewer physical interventions, without policy fine-tuning or failure-specific data.

% We presented Counterfactual Realignment (CoRe), a training-free, model-agnostic framework that lets frozen VLA policies recover from online language switches and physical perturbations. CoRe imagines a counterfactual continuation from a recent viable state, rejoins the policy at a validated pre-contact frame, and restores only the displaced objects that this continuation certifies as necessary, shifting recovery planning from physical trial-and-error to virtual evaluation while preserving completed progress. Across two simulators, four backbones, and a real xArm6, CoRe returns success to near-nominal levels under diverse online disruptions with fewer physical interventions and no policy fine-tuning or failure-specific data.

\appendix

\begin{center}
{\Large\bfseries Technical Appendix}
\end{center}

\setcounter{figure}{0}
\setcounter{table}{0}
\setcounter{equation}{0}
\setcounter{algorithm}{0}

\renewcommand{\thefigure}{A\arabic{figure}}
\renewcommand{\thetable}{A\arabic{table}}
\renewcommand{\theequation}{A\arabic{equation}}
\renewcommand{\thealgorithm}{A\arabic{algorithm}}

\section{CoRe Algorithm}
\label{app:algo}

Algorithm~\ref{alg:core} summarizes one control tick of CoRe around the frozen policy $\pi$. Notation follows the main paper.

\begin{algorithm}
\caption{CoRe control loop (one episode)}
\label{alg:core}
\begin{algorithmic}[1]
\STATE \textbf{loop} at each policy query $t$
\STATE \quad $a_t, h_t \leftarrow$ $K$ stochastic forward passes of $\pi(o_t, \ell')$
\STATE \quad append $(o_t, a_t, q_t, p_t)$ to trace $\mathcal{T}$; $s_t \leftarrow$ Eq.~2 of the main paper
\STATE \quad \textbf{if} $s_t > \tau_k$ for $n_{\text{consec}}$ queries \textbf{then} \textsc{Recover}; \textbf{continue}
\STATE \quad execute chunk $a_t$ open-loop for $N$ control steps
\STATE \textbf{procedure} \textsc{Recover}
\STATE \quad scan $\mathcal{T}$ forward for the divergence onset $d_t > \delta$ (Eq.~3--4)
\STATE \quad refine the anchor within the crossing chunk via $h^{*}$ (Eq.~5); apply phase-aware and clearance gates
\STATE \quad roll out $\tau_B$ from the anchor on synthesized observations (Eq.~6) until $d_{\text{prox}}$; rejoin state $c^{*} \leftarrow$ latest pre-contact frame
\STATE \quad displaced set $\leftarrow$ objects deviating from $c^{*}$ beyond $(\epsilon_{\text{pos}}, \epsilon_{\text{rot}})$; build forced set $\mathcal{F}$
\STATE \quad \textbf{for} candidate subsets $S \supseteq \mathcal{F}$ in increasing size \textbf{do}
\STATE \quad \quad probe $S$ from a hybrid seed; \textbf{if} $d_{\text{arm}}(S) < \epsilon$ (Eq.~7) \textbf{then} $S^{*} \leftarrow S$; \textbf{break}
\STATE \quad \textbf{if} no subset passes \textbf{then} $S^{*} \leftarrow$ all displaced objects
\STATE \quad restore $S^{*}$ (top-first, verified to $\epsilon_{\text{verify}}$, one retry); move arm to $q^{*}$
\STATE \quad truncate $\mathcal{T}$ to the anchor; restore policy RNG; return control to $\pi$
\end{algorithmic}
\end{algorithm}

\section{Benchmark and Dataset Details}
\label{app:data}

\subsection{LangSwitch}
An online language switch is only meaningful if the new instruction is executable from the state the old one has already produced, which requires the skills to share one scene, one object set, and one workspace layout rather than being drawn from separate task suites with their own initial conditions. Existing suites fix a single instruction per episode and place each task in its own scene, so switching between two of them mid-episode either lands the policy in a scene the new instruction does not apply to or silently changes the initial conditions along with the instruction, confounding the switch with a scene change. We therefore build LangSwitch around a single unified tabletop that admits all five skills at once, so any pair of them can be composed into a switch, the pre-switch and post-switch tasks are separated by the instruction alone, and the displaced state the first skill leaves behind is exactly what the second must be recovered from.

The five base skills share one tabletop ManiSkill~\citep{tao2024maniskill3} environment with an xArm6, four colored cubes, two bins, and target markers. The training dataset contains 5{,}000 scripted demonstrations totalling 798{,}571 frames at 20 fps, 1{,}000 episodes per skill, with 401 distinct instruction strings covering color and slot variants. Observations are two RGB cameras (a fixed base camera and a wrist camera, both $512\times512$), and the state and action are the 7-dimensional joint-plus-gripper vector. All backbones are fine-tuned jointly on the same demonstrations.

The nine switch pairs are listed in Table~\ref{tab:pairs}. Target colors are sampled per episode from eight colors with a seed-keyed generator. Re-grounding pairs draw two distinct colors for the same skill. For the push$\to$pull and place$\to$push conflicts the target color is kept and the spatial goal is inverted, so the state produced by the old task contradicts the new goal. The stack$\to$stack conflict instead requests a pyramid with a different top cube, so the partially built structure itself blocks the new goal. The stack$\to$pick pair directs the new skill to the leftover cube not used by the pyramid. Episodes are capped at 600 control steps, and 1{,}200 for stack$\to$stack.

\begin{table}[t]
\centering
\small
\begin{tabular}{lll}
\toprule
Family & Pair & Target relation \\
\midrule
Re-grounding & pick$\to$pick & different color \\
Re-grounding & push$\to$push & different color \\
Re-grounding & place$\to$place & different color and bin \\
Re-skilling & pick$\to$push & different color \\
Re-skilling & place$\to$pull & different color \\
Re-skilling & stack$\to$pick & leftover cube \\
Conflict & push$\to$pull & same color \\
Conflict & place$\to$push & same color \\
Conflict & stack$\to$stack & different top color \\
\bottomrule
\end{tabular}
\caption{The nine LangSwitch pairs. Colors are drawn per episode from eight colors.}
\label{tab:pairs}
\end{table}

\subsection{LIBERO-Long with Perturbation}
We use the official LIBERO-Long suite~\citep{liu2023libero} of 10 long-horizon tasks with the official demonstrations and task definitions, and inject perturbations at evaluation time only. No data is re-collected.

\section{Evaluation Metrics}
\label{app:metrics}

\paragraph{Success.} Success is the simulator's own task predicate, read at every environment step. On LangSwitch it is the ManiSkill success predicate of the post-switch task $\ell'$. On LIBERO it is the official BDDL goal predicate, with a 600-step cap. Episodes where the switch never fires or where $\ell'$ is already satisfied at switch time are invalid and replaced by the next seed.

\paragraph{Number of restorations.} \#Restored counts objects physically returned to their rejoin pose, summed over the recoveries of an episode. A held object that the probe releases in place is put down without being counted. A placement whose verification fails is not counted. For the full-restoration variants, which bypass the probe, the count is reconciled against the displaced set so that held-object bookkeeping matches the probe-based rows.

\paragraph{Post-handoff steps.} For each episode with at least one recovery that ends in success, we count the control steps after the first trigger, excluding the physical steps of the recoveries themselves, then average within each pair and timing cell and macro-average over cells.

\paragraph{Endpoint TCP divergence.} For a synthesized rollout and a ground-truth rollout launched from the same anchor, frames are aligned up to the shorter length and the divergence is the Euclidean distance between tool-center-point positions at the last aligned tick. The same statistic between two independent ground-truth rollouts gives the sampling-variability floor.

\paragraph{Arm-Only weighting.} The Arm-Only variant weights each successful episode by the probability that the arm could have executed its restorations. Each restored object is assigned an orientation bucket from its perturbed pose, upright, tipped, or inverted, and the bucket is mapped to that object's grasp success rate from an oracle suite of 900 motion-planned grasp trials, with 50 seeds per tested object--orientation pair, and orientations whose required grasp width exceeds the gripper aperture are assigned zero as geometrically infeasible. Table~\ref{tab:oracle} lists the full rate table. The per-episode weight is the product over restored objects, and failed episodes contribute zero.

\begin{table}[t]
\centering
\small
\setlength{\tabcolsep}{4pt}
\begin{tabular}{lcc}
\toprule
Object & Upright & Tipped / Inverted \\
\midrule
alphabet soup & 100 & 100 \\
tomato sauce & 100 & 100 \\
butter & 100 & 100 \\
chocolate pudding & 100 & 100 \\
cream cheese & 98 & 100 \\
black book & 100 & 0 \\
akita black bowl & 100 & 100 / 0 \\
porcelain mug & 98 & 64 \\
white--yellow mug & 100 & 64 \\
moka pot & 58 & 36 \\
\bottomrule
\end{tabular}
\caption{Oracle grasp success rates (\%) per object and orientation, exactly as consumed by the Arm-Only weighting. Rates come from 50 motion-planned trials per tested object--orientation pair, 900 trials in total. Non-upright poses share one tested rate per object. The bowl entry lists its side and upside-down rates separately, and its side rate is carried over from the upright grasp rather than measured. The zero entries are orientations whose required grasp width exceeds the gripper aperture and are geometrically infeasible.}
\label{tab:oracle}
\end{table}

\section{Computing Infrastructure}
\label{app:infra}

All checkpoints that we fine-tuned ourselves, covering the three LangSwitch backbones, the LIBERO backbones without official releases, and the real-robot $\pi_{0.5}$, were trained on NVIDIA H200 (141\,GB) GPUs. Officially released LIBERO checkpoints were used where available, including the openpi $\pi_{0.5}$ LIBERO checkpoint and the OpenVLA and OpenVLA-OFT releases. All simulation evaluations ran on two Ubuntu 22.04.3 LTS servers, each with 4$\times$NVIDIA RTX A6000 (48\,GB), one GPU per evaluation shard. LangSwitch used ManiSkill 3.0.1~\citep{tao2024maniskill3} on SAPIEN 3.0.3~\citep{xiang2020sapien}, and LIBERO used robosuite 1.4.0~\citep{zhu2020robosuite} on MuJoCo 3.3.3~\citep{todorov2012mujoco} with the official LIBERO task suite~\citep{liu2023libero}. Policies ran in their own environments, Python 3.10--3.11 with PyTorch 2.2--2.7 or JAX 0.5--0.6 depending on the backbone. Library versions were recorded at evaluation time and are shipped with the code release. The real-robot experiments ran on a separate Linux workstation with two NVIDIA RTX 4090 (24\,GB) cards, one of which serves policy inference while the other hosts the detection and synthesis workers.

\section{Perturbation Injection and Timing}
\label{app:injection}

Both benchmarks inject at a shallow, mid, or deep point of a per-backbone calibrated phase, at fractions 0.30, 0.55, and 0.75 of the phase length.

\subsection{LangSwitch}
The switch tick is $\max(8, \lfloor f \cdot M_A \rfloor)$, where $f$ is the timing fraction and $M_A$ is the backbone's minimum completion time of the pre-switch skill $A$, measured as the earliest success step over 20 clean rollouts per skill (Table~\ref{tab:mincomp}). Evaluation collects 100 valid episodes per cell by scanning seeds upward from 0 within a test band of seeds 0 to 400. An episode is invalid when scene construction fails, when the switch never fires, or when the post-switch goal is already satisfied at switch time, and invalid seeds are skipped for every method identically. Bare and CoRe rows share the same seeds.

\begin{table}[t]
\centering
\small
\begin{tabular}{lccccc}
\toprule
Backbone & pick & push & pull & place & stack \\
\midrule
$\pi_0$ & 59 & 61 & 63 & 102 & 285 \\
$\pi_{0.5}$ & 63 & 55 & 62 & 108 & 299 \\
GR00T N1.7 & 60 & 42 & 61 & 89 & 292 \\
\bottomrule
\end{tabular}
\caption{Per-backbone minimum completion times $M_A$ (control steps) used for switch timing.}
\label{tab:mincomp}
\end{table}

\subsection{LIBERO-Long with Perturbation}
Phase boundaries are calibrated per backbone from 20 clean rollouts per task, taking the minimum success step (full episode), the earliest step at which the target object first moves (pre-grasp phase), and the shortest carrying duration after lift-off (carry phase). Only successful rollouts enter the calibration. Injection parameters follow.

\paragraph{Arm Collision.} A random torque impulse is applied to all seven arm joints for 5 environment steps, drawn as a standard normal direction scaled to magnitude 50. An injection is accepted only if no movable object drifts more than 0.15\,m in the plane or rotates more than 20$^\circ$, with up to 8 redraws from a state snapshot.

\paragraph{Object Tip.} A horizontal-axis torque wrench acts on the target for 4 steps. The magnitude is 0.05\,N$\cdot$m scaled by the square root of the object's relative mass, with per-object attenuation for four light objects tuned from visual inspection. The injection is rejected if any neighbor moves more than 0.05\,m or 20$^\circ$. After a 20-step settle, episodes whose target leaves a 0.75\,m workspace radius are invalid.

\paragraph{Object Displace.} The target is relocated 0.12\,m from its current position along a uniformly sampled planar direction, with up to 64 angle attempts. A candidate is rejected if it contacts any non-table geometry or leaves the 0.75\,m reach radius. If the target is in the gripper the injection is deferred to the next tick.

\paragraph{Grasp Failure.} The first environment step at which the target rises 1\,cm above its spawn height anchors the carry phase, and at the scheduled tick both gripper fingers are opened in place, dropping the object. Episodes where the open gripper still penetrates the target are invalid.

\paragraph{Guards.} For the two pre-contact families, if the tool-center-point closes within 6\,cm of the target surface before the scheduled tick, the injection fires immediately so that the perturbation always precedes contact. Each cell collects 20 valid episodes by scanning seeds from 0, with invalid episodes replaced by the next seed.

\section{Backbone-Specific Handoff}
\label{app:handoff}

\paragraph{Common steps.} The rejoin frame is the latest pre-interaction frame of $\tau_B$, and the gripper target is taken from that frame. After restoration and the arm move, the pending instruction is consumed, observations are refreshed, the policy's sampling RNG is restored to its value at recovery entry, and the trace is truncated to the anchor so that later recoveries chain from the current one.

\paragraph{Chunked delta backbones.} $\pi_0$, $\pi_{0.5}$, and GR00T predict delta actions from stateless adapters, so no proprioceptive state needs re-anchoring. When the imagined rollout truncates mid-chunk after $h^{*}$ safe control steps, only the unexecuted remainder of that chunk is handed back for execution, which prevents the first $h^{*}$ steps from running twice.

\paragraph{X-VLA.} X-VLA emits 30-step absolute end-effector chunks and maintains an open-loop proprioceptive state, which CoRe saves and re-anchors at every boundary between real and imagined execution, and re-initializes from the controller pose after realignment. Because a full chunk runs without re-observation, the rejoin frame must additionally keep at least 0.15\,m of surface clearance from every movable object. Before handoff the retained chunk is cut at its first gripper-close command, the realigned arm pose is polished toward the first retained waypoint, and execution rejoins the chunk at the waypoint closest to the realigned tool-center-point.

\paragraph{Benchmark split.} On LangSwitch every restoration is executed physically, with put-down for held objects whose position the probe deems irrelevant, motion-planned pick-and-place otherwise, and a three-segment arm move with a 20-step gripper ramp. On LIBERO restorations are applied by direct state writes to object poses, articulated joints, and the arm configuration, and the retained chunk remainder is then executed normally.

\section{CoRe Hyperparameters and Calibration}
\label{app:hyper}

Table~\ref{tab:hyper} lists every CoRe hyperparameter with the value used in all reported experiments. The anchor threshold $\delta$ differs between benchmarks because the divergence lives in different spaces. On LangSwitch the instruction changes, so $d_t$ compares std-normalized actions and $\delta$ is dimensionless. On LIBERO the instruction is unchanged, and because the arm there is kinematically redundant, the configuration deviation of Eq.~4 of the main paper is operationalized in task space: the clean and recorded arm configurations are mapped through forward kinematics and $d_t$ is the Euclidean distance between the induced tool-center-point positions, so $\delta$ carries meters. The task-space form is used because nullspace drift of the redundant arm raises the joint-space noise floor to the same order as the perturbation signal, whereas the two remain separable at the end effector. This operationalization applies only to the unchanged-instruction LIBERO case; LangSwitch instead uses the standardized action divergence defined for $\ell \neq \ell'$ in Eq.~4 of the main paper. The subset threshold $\epsilon$ is a mean joint-space distance and transfers across the two arms unchanged.

\begin{table}[t]
\centering
\small
\setlength{\tabcolsep}{3.5pt}
\begin{tabular}{lll}
\toprule
Parameter & Value & Notes \\
\midrule
Detection samples $K$ & 8 & mean-pooled \\
Trigger $n_{\text{consec}}$ & 3 & 3-of-3 raw scores \\
Plateau backup & 5 @ $0.9\tau_k$ & sustained scores \\
Refractory period & 16 queries & between triggers \\
$\tau_k$ quantile & 0.95 & held-out successes \\
Bank size (LS) & 50 eps/task & successes only \\
Covariance ridge & $10^{-2}$ & Eq.~2 \\
Anchor $\delta$ (LS) & 0.09 & normalized actions \\
Anchor $\delta$ (LB) & 0.08\,m & TCP deviation \\
Rollout $d_{\text{prox}}$ (LS) & 0.02\,m & object surface \\
Rollout $d_{\text{prox}}$ (LB) & 0.08\,m & OBB surface \\
Rollout cap & 15 ticks & wall-clock 10\,s \\
Anchor clearance (LB) & 0.08\,m & TCP to objects \\
X-VLA standoff & 0.15\,m & rejoin clearance \\
Displaced $\epsilon_{\text{pos}}$ (xy) & 0.03\,m & vs. rejoin pose \\
Displaced $\epsilon_{\text{pos}}$ (z) & 0.005\,m & vs. rejoin pose \\
Displaced $\epsilon_{\text{rot}}$ & 0.30\,rad & quaternion angle \\
Forced-set $r_{\text{veto}}$ & 0.08\,m & task entities \\
Goal-region radius & 0.06\,m & target area \\
Subset $\epsilon$ & 0.050 & mean joint L2 \\
Verify $\epsilon_{\text{verify}}$ & 0.03\,m & one retry \\
Restore cap & 1 per object & per episode \\
\bottomrule
\end{tabular}
\caption{CoRe hyperparameters. LS denotes LangSwitch and LB denotes LIBERO-Long.}
\label{tab:hyper}
\end{table}

\paragraph{Threshold search.} Both swept thresholds were tuned on a validation seed band, seeds 500 to 800, disjoint from the test band, on three validation pairs covering the three switch families with 90 episodes per candidate. Candidates for $\delta$ were the empirical q10 to q90 quantiles $\{0.05, 0.09, 0.14, 0.18, 0.24\}$ of scouted divergence values. Candidates for $\epsilon$ were nine deciles $\{0.008$ to $0.060\}$ of the arm-divergence distribution scouted with an accept-everything probe. Selection used the same rule for both. Candidates whose success rate lies within 3 points of the best form the admissible set, and within it the candidate minimizing physical cost is chosen, giving $\delta^{*}=0.09$ and $\epsilon^{*}=0.050$ (Figs.~\ref{fig:delta_sweep} and \ref{fig:eps_sweep}).

\paragraph{Seeds.} Test episodes scan seeds 0 to 400. Manifest-based runs additionally seed the policy RNG per episode as $10^6$ plus the environment seed. Detection banks are collected from seeds 0 to 500 with a fixed 70/30 train and calibration split.

\section{Observation Synthesis Details}
\label{app:synth}

\paragraph{Voxel stores.} Per-object stores use 2\,mm voxels in each object's local frame, accumulated incrementally from segmented RGB-D frames with first-seen colors, using the simulator's instance segmentation to separate arm, objects, and background. Capture runs asynchronously so that fusion never blocks the control loop. A world-frame background store accumulates all non-arm non-object pixels, and pixels with invalid depth are kept on a zero-parallax plate that fills regions the background store cannot cover.

\paragraph{Exocentric view.} The anchor frame is edited in place. The removal mask is the union of the arm silhouette and the anchor-pose voxel footprints of displaced objects, which also covers pixels the segmentation misses under arm occlusion. Vacated pixels are filled by projecting the background store, then the plate, and residual holes are inpainted. Pixels of undisplaced objects that the fill overwrote are repaired by reprojecting their stores at the anchor pose. Displaced objects are then splatted at their imagined poses with z-buffering, and the arm is composited last from the forward-kinematic render.

\paragraph{Egocentric view.} The wrist viewpoint moves with the imagined arm, so the entire image is reprojected from the stores at the forward-kinematic camera pose, background first with caching, then objects with z-buffered layering, followed by despeckling of sparse pixels, plate filling, and inpainting.

\paragraph{Inpainting.} Holes covering at least 10\% of the image use LaMa when available, and all other holes use OpenCV Telea inpainting. Splatting uses a deterministic GPU z-buffer that resolves duplicate writes by depth ordering, so repeated synthesis of the same state is bit-stable.

\paragraph{Arm rendering.} The arm is rendered by a pinocchio forward-kinematics chain and pyrender in a worker process. On LIBERO the render lighting is calibrated to the MuJoCo headlight, a 0.03 ambient term plus a camera-following directional light, since the uncalibrated render composited the arm 2 to 3 times brighter than the ground-truth frames. On LangSwitch the rendered arm already matches the SAPIEN ground truth closely, so no lighting calibration is applied. Imagined proprioception is kept consistent with the imagined pose by overriding joint positions from forward kinematics with zero velocities, together with the derived end-effector pose on LIBERO.

\paragraph{Failure handling.} A synthesis failure inside a probe aborts that probe rather than silently substituting ground truth. If all probes fail, CoRe falls back to restoring every displaced object, so synthesis failures can only reduce restoration efficiency rather than cause the probe to accept an insufficient restoration subset.

\subsection{Qualitative Examples}

\begin{figure}[t]
    \centering
    \includegraphics[width=\linewidth]{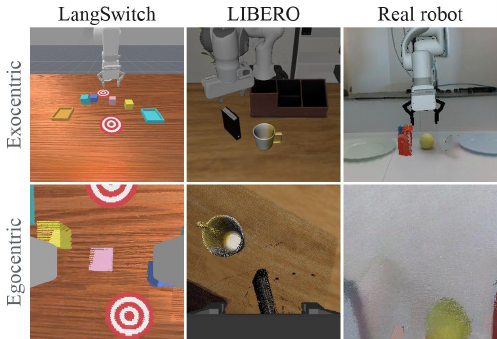}
    \caption{\textbf{Synthesized observations, as consumed by the frozen policy.} Columns are the three domains, rows are the two viewpoints. LangSwitch shows a frame of an imagined continuation, in which the target is rendered in mid-air because the counterfactual has already grasped it. LIBERO shows the dual-view synthesis of a single imagined state. The real-robot column shows the repaint a subset probe is evaluated on, with the candidate objects removed and the vacated pixels filled from the background store.}
    \label{fig:synth_examples}
\end{figure}

Figure~\ref{fig:synth_examples} shows synthesized observations exactly as the frozen policy receives them, for the three domains and both viewpoints. Each column comes from the same pipeline; only the calibration, the segmentation source, and the render lighting differ.

On LangSwitch the frame is taken from an imagined continuation $\tau_B$. The target cube is splatted in mid-air because the counterfactual has already lifted it: object poses in an imagined rollout follow the imagined arm, so a grasped object is rendered where the imagined gripper holds it rather than where it still rests in the recorded scene. The egocentric view shows the same state from the moving wrist camera, with the arm composited last from the forward-kinematic render.

On LIBERO the pair illustrates the dual-view case: the exocentric and the egocentric stream are both re-synthesized for the same imagined state. LIBERO recoveries are dominated by re-approach rather than by scene restoration, so these frames exercise the synthesis path alone, without a subset probe.

On the real robot the frame is the repaint that a subset probe consumes. The candidate objects are removed from the anchor frame, the background store and the zero-parallax plate fill the vacated pixels, and the remaining objects are reprojected at their anchor poses, so the policy is asked what it would do if that subset had been restored. Residual speckle at object boundaries is the visible cost of a real depth sensor, and it is the input on which the probe decision is actually made.

Lighting is the one part of the composite that must be calibrated per domain, because the arm is rendered by an external forward-kinematic renderer rather than by the simulator that produced the background. On LIBERO the render is matched to the MuJoCo headlight with a $0.03$ ambient term plus a camera-following directional light; without this correction the composited arm is $2$ to $3$ times brighter than the ground-truth frames, and that mismatch alone is a distribution shift the policy can key on. On LangSwitch the rendered arm already agrees with the SAPIEN ground truth, so no correction is applied. On the real robot no synthetic arm is composited at all, and the room illumination in the retained pixels is whatever the scene had, so the question does not arise. The figure is deliberately shown without any per-panel exposure adjustment, so the brightness agreement between the rendered arm and its surrounding scene is what the policy actually sees.

\section{Ablation Protocols}
\label{app:ablation}

\subsection{Fidelity of Synthesized Rollouts}
On LangSwitch the protocol covers all 9 pairs at 3 timings with 10 episodes each per backbone, yielding 278, 281, and 305 recovery anchors for $\pi_0$, $\pi_{0.5}$, and GR00T. On LIBERO it covers all 10 tasks, 4 perturbation types, and 3 timings with one episode each, yielding 137, 96, and 130 anchors for $\pi_{0.5}$, X-VLA, and GR00T. At every anchor three independent closed-loop rollouts are launched, one on synthesized observations and two on ground-truth renders, all sharing the benchmark's contact-boundary threshold. The synth-to-GT endpoint divergence is reported against the GT-to-GT divergence as the sampling-variability floor. On LIBERO this floor is structurally near zero, so absolute values are reported there.

\subsection{Recovery Mechanism Ablation}
All four variants replay one fixed manifest of 270 episodes, 9 pairs at 3 timings with 10 episodes, on $\pi_{0.5}$ with identical environment seeds, policy seeds, and switch ticks. The variants differ only in the realignment target. Episode Reset restores the initial state, Anchor Rollback restores everything at the recovery anchor, CoRe w/o Sub follows the counterfactual continuation with full restoration, and CoRe uses the probe-selected subset with $\epsilon = 0.050$. The three full-restoration variants do not consume $\epsilon$.

\subsection{Threshold Sensitivity}
\begin{figure}[t]
    \centering
    \includegraphics[width=\linewidth]{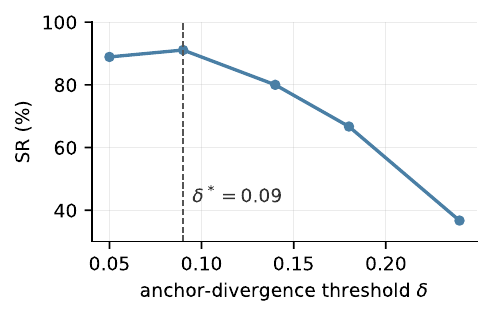}
    \caption{Anchor-divergence threshold sweep on the validation pairs. Success peaks at $\delta=0.09$ and collapses for later anchors.}
    \label{fig:delta_sweep}
\end{figure}

\begin{figure}[t]
    \centering
    \includegraphics[width=\linewidth]{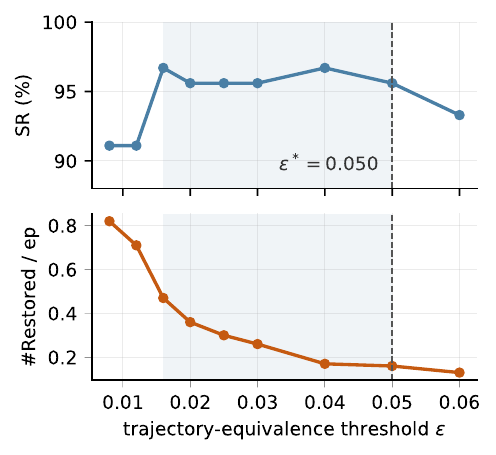}
    \caption{Trajectory-equivalence threshold sweep on the validation pairs. The shaded band is the admissible plateau, and $\epsilon^*=0.050$ minimizes restorations within it.}
    \label{fig:eps_sweep}
\end{figure}

Figure~\ref{fig:delta_sweep} shows the $\delta$ sweep on the validation pairs, 90 episodes per candidate. Success peaks at $\delta = 0.09$ with 91.1\% and degrades monotonically for later anchors, reaching 36.7\% at $\delta = 0.24$, since an anchor placed too deep resumes the policy inside the lock-in regime discussed in the main paper. Figure~\ref{fig:eps_sweep} shows the $\epsilon$ sweep. Overly tight thresholds force near-full restoration and lose 4 to 5 points to the added physical failure surface, a broad plateau spans 0.016 to 0.050, and the loosest setting begins to skip required restorations. Within the plateau the largest $\epsilon$ minimizes restorations at 0.16 objects per episode.

\section{Additional Results}
\label{app:results}

Tables~\ref{tab:langswitch_full} and \ref{tab:libero_full} report the per-cell success rates behind the switch columns of Table 1 and behind Table 2 of the main paper, and Table~\ref{tab:noswitch} reports the per-skill No-Switch success rates behind the No-Switch and All columns of Table 1. Every LangSwitch cell aggregates 100 episodes and every LIBERO entry aggregates 200 episodes, with cell counts verified against the raw logs. The 1{,}500 unswitched CoRe episodes of Table~\ref{tab:noswitch} contain the zero false triggers reported in the main paper. The No-Switch rows share the same environment seed sets, while policy sampling remains stochastic and the detector draws $K$ samples at every query, so the wrapped execution is not bit-identical to a bare rollout. The small per-skill differences in Table~\ref{tab:noswitch} therefore reflect sampling noise rather than any intervention, as no recovery is ever triggered.

\begin{table*}[!t]
\centering
\scriptsize
\setlength{\tabcolsep}{5pt}
\renewcommand{\arraystretch}{0.95}
\begin{tabular}{llrrrrrr}
\toprule
& & \multicolumn{3}{c}{Bare} & \multicolumn{3}{c}{+CoRe} \\
\cmidrule(lr){3-5}\cmidrule(lr){6-8}
Family & Pair & S & M & D & S & M & D \\
\midrule
\multicolumn{8}{l}{$\pi_0$} \\
Re-grounding & pick$\to$pick & 24.0 & 1.0 & 0.0 & 99.0 & 97.0 & 100.0 \\
Re-grounding & push$\to$push & 1.0 & 0.0 & 1.0 & 86.0 & 89.0 & 88.0 \\
Re-grounding & place$\to$place & 0.0 & 0.0 & 0.0 & 74.0 & 75.0 & 70.0 \\
Re-skilling & pick$\to$push & 18.0 & 2.0 & 3.0 & 96.0 & 96.0 & 95.0 \\
Re-skilling & place$\to$pull & 5.0 & 12.0 & 20.0 & 87.0 & 87.0 & 92.0 \\
Re-skilling & stack$\to$pick & 41.0 & 3.0 & 37.0 & 97.0 & 99.0 & 90.0 \\
Conflict & push$\to$pull & 2.0 & 0.0 & 0.0 & 74.0 & 89.0 & 87.0 \\
Conflict & place$\to$push & 100.0 & 89.0 & 84.0 & 97.0 & 99.0 & 95.0 \\
Conflict & stack$\to$stack & 31.0 & 10.0 & 10.0 & 77.0 & 62.0 & 65.0 \\
\addlinespace[2pt]
\multicolumn{2}{l}{Mean (per timing)} & 24.7 & 13.0 & 17.2 & 87.4 & 88.1 & 86.9 \\
\midrule
\multicolumn{8}{l}{$\pi_{0.5}$} \\
Re-grounding & pick$\to$pick & 87.0 & 0.0 & 0.0 & 98.0 & 98.0 & 97.0 \\
Re-grounding & push$\to$push & 64.0 & 0.0 & 0.0 & 100.0 & 97.0 & 98.0 \\
Re-grounding & place$\to$place & 0.0 & 0.0 & 0.0 & 88.0 & 86.0 & 87.0 \\
Re-skilling & pick$\to$push & 79.0 & 9.0 & 11.0 & 100.0 & 99.0 & 100.0 \\
Re-skilling & place$\to$pull & 51.0 & 62.0 & 56.0 & 97.0 & 96.0 & 95.0 \\
Re-skilling & stack$\to$pick & 98.0 & 0.0 & 99.0 & 93.0 & 94.0 & 91.0 \\
Conflict & push$\to$pull & 21.0 & 0.0 & 0.0 & 84.0 & 98.0 & 95.0 \\
Conflict & place$\to$push & 94.0 & 89.0 & 72.0 & 100.0 & 99.0 & 97.0 \\
Conflict & stack$\to$stack & 42.0 & 11.0 & 14.0 & 85.0 & 69.0 & 62.0 \\
\addlinespace[2pt]
\multicolumn{2}{l}{Mean (per timing)} & 59.6 & 19.0 & 28.0 & 93.9 & 92.9 & 91.3 \\
\midrule
\multicolumn{8}{l}{GR00T N1.7} \\
Re-grounding & pick$\to$pick & 6.0 & 7.0 & 2.0 & 97.0 & 94.0 & 90.0 \\
Re-grounding & push$\to$push & 10.0 & 2.0 & 0.0 & 80.0 & 71.0 & 77.0 \\
Re-grounding & place$\to$place & 0.0 & 0.0 & 0.0 & 78.0 & 73.0 & 73.0 \\
Re-skilling & pick$\to$push & 11.0 & 3.0 & 0.0 & 85.0 & 95.0 & 89.0 \\
Re-skilling & place$\to$pull & 27.0 & 29.0 & 27.0 & 91.0 & 91.0 & 86.0 \\
Re-skilling & stack$\to$pick & 19.0 & 0.0 & 19.0 & 92.0 & 94.0 & 77.0 \\
Conflict & push$\to$pull & 22.0 & 1.0 & 0.0 & 85.0 & 82.0 & 83.0 \\
Conflict & place$\to$push & 85.0 & 93.0 & 88.0 & 89.0 & 94.0 & 92.0 \\
Conflict & stack$\to$stack & 35.0 & 8.0 & 8.0 & 64.0 & 53.0 & 56.0 \\
\addlinespace[2pt]
\multicolumn{2}{l}{Mean (per timing)} & 23.9 & 15.9 & 16.0 & 84.6 & 83.0 & 80.3 \\
\bottomrule
\end{tabular}
\caption{LangSwitch per-cell success rates (\%). Each cell is 100 episodes. S/M/D denote shallow/mid/deep switch timing.}
\label{tab:langswitch_full}
\end{table*}

\begin{table*}[!t]
\centering
\scriptsize
\setlength{\tabcolsep}{3.5pt}
\renewcommand{\arraystretch}{0.95}
\begin{tabular}{lrrrrrrrrrrrrl}
\toprule
& \multicolumn{3}{c}{Arm collision} & \multicolumn{3}{c}{Grasp failure} & \multicolumn{3}{c}{Object displace} & \multicolumn{3}{c}{Object tip} & \\
\cmidrule(lr){2-4}\cmidrule(lr){5-7}\cmidrule(lr){8-10}\cmidrule(lr){11-13}
Method & S & M & D & S & M & D & S & M & D & S & M & D & Avg \\
\midrule
OpenVLA & 1.0 & 3.0 & 6.0 & 42.5 & 42.0 & 39.5 & 4.0 & 3.5 & 2.0 & 3.0 & 4.0 & 3.0 & 12.8$\pm$0.3 \\
RobustVLA & 10.0 & 10.5 & 19.0 & 34.0 & 50.5 & 57.0 & 3.0 & 1.5 & 3.5 & 2.5 & 3.5 & 4.0 & 16.6$\pm$4.3 \\
GR00T N1.7 & 8.5 & 18.0 & 20.5 & 40.5 & 49.5 & 48.0 & 19.5 & 17.0 & 12.5 & 13.5 & 14.5 & 18.5 & 23.4$\pm$2.5 \\
OpenVLA-OFT & 15.5 & 17.0 & 23.5 & 54.5 & 67.5 & 73.0 & 20.5 & 12.0 & 8.0 & 13.5 & 6.0 & 11.5 & 26.9$\pm$1.8 \\
VLA-Adapter & 15.0 & 22.5 & 26.5 & 60.5 & 65.0 & 69.5 & 13.0 & 10.0 & 9.5 & 11.0 & 7.5 & 13.0 & 26.9$\pm$2.4 \\
X-VLA & 57.0 & 66.5 & 58.0 & 44.5 & 59.0 & 50.0 & 6.5 & 5.5 & 6.5 & 10.5 & 9.0 & 9.5 & 31.9$\pm$2.8 \\
$\pi_{0.5}$ & 35.0 & 29.5 & 46.5 & 66.0 & 66.0 & 63.0 & 51.0 & 43.5 & 37.0 & 30.5 & 25.0 & 28.5 & 43.5$\pm$2.3 \\
\midrule
GR00T N1.7 + CoRe & 80.0 & 78.5 & 64.5 & 71.0 & 72.0 & 71.5 & 68.5 & 77.0 & 71.0 & 46.5 & 64.5 & 72.5 & 69.8$\pm$3.3 \\
X-VLA + CoRe & 85.0 & 88.5 & 81.0 & 80.5 & 83.0 & 81.5 & 91.5 & 90.0 & 83.5 & 89.0 & 80.0 & 80.0 & 84.5$\pm$2.6 \\
$\pi_{0.5}$ + CoRe & 92.0 & 86.5 & 88.5 & 87.0 & 93.0 & 90.5 & 86.5 & 88.5 & 87.5 & 84.0 & 88.0 & 87.0 & 88.2$\pm$0.8 \\
\bottomrule
\end{tabular}
\caption{LIBERO-Long per-family per-timing success rates (\%). Each entry aggregates 200 episodes. S/M/D denote shallow/mid/deep perturbation timing, and Avg reports mean and sample standard deviation across the three timings.}
\label{tab:libero_full}
\end{table*}

\begin{table}[t]
\centering
\small
\setlength{\tabcolsep}{4pt}
\begin{tabular}{lcccccc}
\toprule
& \multicolumn{2}{c}{$\pi_0$} & \multicolumn{2}{c}{$\pi_{0.5}$} & \multicolumn{2}{c}{GR00T N1.7} \\
\cmidrule(lr){2-3}\cmidrule(lr){4-5}\cmidrule(lr){6-7}
Skill & Bare & +CoRe & Bare & +CoRe & Bare & +CoRe \\
\midrule
pick & 100.0 & 100.0 & 98.0 & 100.0 & 99.0 & 100.0 \\
push & 94.0 & 97.0 & 100.0 & 100.0 & 91.0 & 92.0 \\
place & 90.0 & 91.0 & 97.0 & 93.0 & 81.0 & 89.0 \\
pull & 93.0 & 91.0 & 98.0 & 99.0 & 95.0 & 94.0 \\
stack & 76.0 & 71.0 & 96.0 & 94.0 & 76.0 & 76.0 \\
\addlinespace[2pt]
Mean & 90.6 & 90.0 & 97.8 & 97.2 & 88.4 & 90.2 \\
\bottomrule
\end{tabular}
\caption{No-Switch per-skill success rates (\%) on LangSwitch. Each cell aggregates 100 episodes.}
\label{tab:noswitch}
\end{table}

\section{Real-World System Details}
\label{app:real}

\subsection{Rig and Perception}

The platform is a 6-DoF UFACTORY xArm6 with a parallel-jaw gripper on a fixed tabletop, commanded in joint space at $10$\,Hz, which matches the demonstration recording rate and the LeRobot dataset frame rate, so training and deployment share one cadence. Two Intel RealSense D435i cameras provide RGB-D: one fixed third-person camera viewing the workspace from the front-left and one wrist camera rigidly mounted on the gripper flange. Both streams are captured at $640\times480$ at $30$\,fps and center-cropped to $512\times512$ for the policy, with the intrinsics shifted by the crop at the source so that every downstream consumer uses one camera matrix. Depth is aligned to color in hardware; readings outside $(0, 3]$\,m are treated as invalid.

Two extrinsics are calibrated offline and then frozen: the third-person camera-to-base transform and the wrist camera hand-eye transform. The third-person calibration is verified by commanding the tool to a set of known poses and reprojecting, with a residual of $4.3$\,mm at the tool center point. A table-height calibration fixes the support plane at $z=-0.183$\,m in the base frame; all vertical safety rails are expressed relative to it.

Perception replaces the simulator's privileged state with three components. Object detection uses an open-vocabulary detector (Grounding-DINO) with a per-object text prompt, guarded against cross-prompt ambiguity so that a box claimed with higher confidence by another object's prompt is rejected. Instance masks come from a promptable segmentation model (SAM2) seeded by those boxes, with an HSV-band segmenter as a fallback. The arm mask, which the simulator supplies for free, is produced by rendering the arm from the measured joint configuration through a forward-kinematics chain and dilating the silhouette by $9$\,px to absorb calibration slop. Object poses use shape-specific estimators: yaw-only pose from the depth footprint for boxes, and analytic geometry for the sphere and cylinder, whose curved tops lose stereo depth at this grazing viewing angle.

Observation synthesis reuses the simulator pipeline unchanged in structure: $2$\,mm voxels, first-seen colors, a world-frame background store and per-object stores in object-local frames, a zero-parallax plate for depth-invalid pixels, and inpainting that switches to LaMa when a hole covers at least $10\%$ of the frame. Three differences from simulation are forced by real sensing. Voxel capture runs asynchronously off the control path, since a real RGB-D fusion step cannot block a $10$\,Hz loop. Per-voxel view directions are retained so that render-time backface culling can suppress voxels observed only from grazing angles. Synthesis renders at the policy input resolution rather than the native camera resolution, because the backbone resizes to $224\times224$ regardless. As in simulation, a synthesis failure aborts the affected probe rather than silently substituting a live observation.

\subsection{Data Collection and Fine-Tuning}

We collect $100$ demonstrations for each of the three skills, pick, place, and push. Episodes that fail a quality gate are discarded, and recording is restricted to the manipulation segment to exclude idle motion and trailing holds, so the resulting dataset contains only clean, unperturbed executions. Episodes cycle round-robin over three household objects, a tennis ball, a red box, and a blue gum bottle, with two plates serving as placement targets.

We fine-tune $\pi_{0.5}$ on this data with the official JAX pipeline. The dataset is converted to LeRobot v2.1 with the standard image, wrist image, state, and action features, and normalization statistics are computed on the converted set. Actions are absolute joint positions with a delta transform; the model emits $10$-step chunks of which $8$ are executed before the next query, matching the deployed control cadence. The model is fully fine-tuned, all parameters included, on a single NVIDIA H200 GPU, and the workstation's RTX 4090 cards serve inference only at deployment. Only static, unperturbed demonstrations enter training. The policy never sees a language switch, a collision, or a displaced object during fine-tuning, so every real-robot result is out of the training distribution by construction.

\subsection{Perturbation Execution}

Real evaluation covers five conditions: the three LangSwitch families instantiated on this rig, an Arm Collision family, and a Scene Rearrangement family. Each condition contributes $20$ episodes split across shallow, mid, and deep injection timings in the same $0.30 / 0.55 / 0.75$ proportions used in simulation. The phase boundaries are recalibrated on the real robot rather than inherited: for each base skill we take the earliest step at which the nominal policy satisfies the task over clean rollouts on this rig, and derive the three injection ticks from that minimum completion time. Timings are stratified over seeds so that no timing cell is confounded with a particular layout.

\paragraph{Arm Collision.} The collision is applied in joint space as a twist injected at the scheduled tick. Every joint participates: each joint's magnitude is drawn uniformly from $[0.12, 0.32]$\,rad with a random sign, then scaled by fixed per-joint weights that down-weight the base joint, whose sweep dominates tool swing and would push the arm out of the workspace, and the two wrist joints, which sit near singularity. A candidate twist is accepted only if forward kinematics places the resulting tool pose at least $0.10$\,m above the table, inside the workspace bounds with a $30$\,mm margin, and clear of joint limits; rejected draws are re-sampled with a gentle magnitude shrink. Because the impulse is a sampled quantity, it is recorded per episode and replayed verbatim on the paired baseline episode, so the two arms of a pair experience the same physical disturbance.

\paragraph{Scene Rearrangement.} This condition is executed by the operator rather than by code, and it is deliberately harder than its simulated counterpart. At the scheduled tick the controller freezes the arm on its last commanded target and prompts the operator, who displaces \emph{all three} objects on the table, not only the instruction target, and then releases the gate. The controller drops any stale internal object registry, re-observes the scene, and returns control to the policy. Restoring everything is therefore never the cheap answer: the recovery has to decide which of the moved objects actually matter for the task. Object identities and coarse layout are held fixed within a pair so that the baseline and CoRe episodes face the same rearrangement.

\paragraph{Pairing.} Baseline and CoRe episodes are paired by initial layout, instruction, and, for Arm Collision, by the replayed impulse. Pairs are run back to back so that lighting and calibration drift cannot separate the two arms.

\subsection{Recovery Execution and Measurement}

\paragraph{Physical restoration.} Unlike the simulator, which can write object poses directly, every real restoration is executed physically. A held object whose position the probe deems irrelevant is put down in place rather than carried. Objects that must be restored are picked and placed by a motion-planned three-segment move, approach, transport at a fixed carry height, and descent, with a ramped gripper close and a per-object closing target taken from the object registry. Placement is verified by re-observing the object after release; a placement whose verification fails is retried once and, if it fails again, is not counted as restored. The arm is then moved to the rejoin configuration, the pending instruction is consumed, observations are refreshed, and the policy's sampling RNG is restored to its value at recovery entry, exactly as in simulation.

\paragraph{Safety rails.} Every commanded step passes through the same rails in both arms: a tool-center-point floor $3$\,mm above the calibrated table, a per-step joint delta clamp, a workspace bound check, and a Cartesian speed cap. A controller-side speed fault aborts the episode rather than being retried silently. These rails are identical for the baseline and for CoRe, so they cannot advantage either arm.

\paragraph{Timing.} Two clocks are recorded per recovery. Planning time runs from the moment the trigger is adjudicated to the moment the realignment target is committed, and therefore contains the anchor scan, the imagined rollout, and the subset probes. Execution time runs from that commit to the end of the physical realignment, and therefore contains only arm and object motion. The reported medians are $9.7$\,s of planning and $8.8$\,s of physical motion per recovery over $165$ recoveries in $100$ layout-paired episodes. The two are dominated by different regimes: planning is essentially the imagined rollout, about $1.2$\,s per imagined chunk at a median of $8$ chunks, while the median recovery restores no object at all and spends its physical budget realigning the arm alone. Recoveries that do carry an object are far more expensive, which is why a single manipulation is reported separately at $20.5$\,s of motion and $46.6$\,s of wall clock.

\paragraph{Safety adjudication.} Collisions and emergency stops are logged by the controller with the stop reason attached to the episode record and are confirmed by the operator at episode end, so a physically unsafe episode cannot be scored as a success by a kinematic heuristic. Under this accounting $21$ of the $100$ baseline episodes end in a collision or an emergency stop, which is $27\%$ of all baseline failures, against $5$ for CoRe, one of which is the injected impulse itself striking clutter before any recovery could be triggered.

\subsection{Real-Robot Results in Detail}

\begin{figure}[t]
    \centering
    \includegraphics[width=\linewidth]{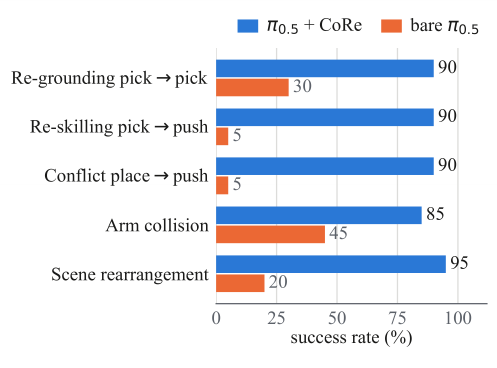}
    \caption{Real-robot success rate per condition, $20$ episodes each, for the bare policy and for the same frozen policy wrapped in CoRe.}
    \label{fig:real_success}
\end{figure}

Figure~\ref{fig:real_success} gives the per-condition success rates behind the real-robot table of the main paper: the frozen policy alone succeeds on $21$ of $100$ perturbed episodes, and the same policy wrapped in CoRe succeeds on $90$, with the largest margins on the two conditions whose pre-perturbation behaviour is furthest from the post-perturbation goal.

\begin{figure}[!tb]
    \centering
    \includegraphics[width=0.8\linewidth]{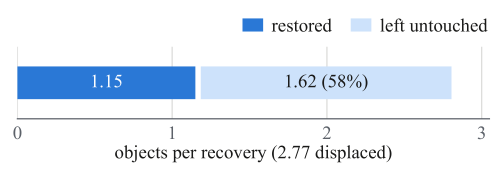}
    \caption{Recovery composition under Scene Rearrangement: of the objects displaced above tolerance, the share the probe restores and the share it declines to touch.}
    \label{fig:real_subset}
\end{figure}

Figure~\ref{fig:real_subset} shows what the subset probe buys. In a typical Scene Rearrangement recovery $2.77$ objects sit beyond tolerance, yet only $1.15$ are physically restored: $58\%$ of the displaced objects are deliberately left where the operator moved them, because the probe finds their displacement irrelevant to the pending task.

\begin{table}[!tb]
\centering
\small
\begin{tabular}{lcccc}
\toprule
 & Bare & \multicolumn{3}{c}{+CoRe} \\
\cmidrule(lr){3-5}
Condition & all & S & M & D \\
\midrule
Re-grounding pick$\to$pick & 6/20 & 7/7 & 5/7 & 6/6 \\
Re-skilling pick$\to$push & 1/20 & 6/7 & 6/7 & 6/6 \\
Conflict place$\to$push & 1/20 & 6/7 & 7/7 & 5/6 \\
Arm collision & 9/20 & 6/7 & 5/7 & 6/6 \\
Scene rearrangement & 4/20 & 7/7 & 6/7 & 6/6 \\
\midrule
Total & 21/100 & 32/35 & 29/35 & 29/30 \\
\bottomrule
\end{tabular}
\caption{Real-robot success by injection timing. S/M/D denote shallow/mid/deep, at the same $0.30/0.55/0.75$ fractions of the recalibrated real phase.}
\label{tab:real_timing}
\end{table}

Table~\ref{tab:real_timing} breaks the CoRe column down by injection timing. Success does not decay as the perturbation is injected deeper into the skill, which is the behaviour the anchor search is designed to produce: a later perturbation moves the anchor later as well, so the imagined continuation still starts before the policy commits to the stale plan.

\begin{table}[!tb]
\centering
\small
\begin{tabular}{lcc}
\toprule
 & Bare & +CoRe \\
\midrule
Collisions or emergency stops & 21/100 & 5/100 \\
\bottomrule
\end{tabular}
\caption{Physical safety over the $100$ layout-paired episodes. CoRe removes three quarters of the unsafe terminations in absolute count.}
\label{tab:real_safety}
\end{table}

Table~\ref{tab:real_safety} reports the physical-safety accounting, which is the reading a deployment cares about most. A stale plan executed after a perturbation does not merely fail the task: on this rig it drives the arm into the table or into an object, and $21$ of the $100$ baseline episodes end in a collision or an operator emergency stop, $27\%$ of all baseline failures. Under CoRe the count falls to $5$, one of which is the injected impulse itself striking clutter before any recovery could be triggered.

\begin{table}[!tb]
\centering
\small
\begin{tabular}{lccc}
\toprule
Skill & pick & place & push \\
\midrule
Nominal success & 19/20 & 19/20 & 20/20 \\
\bottomrule
\end{tabular}
\caption{Unperturbed control condition, frozen checkpoint, $20$ episodes per skill.}
\label{tab:real_nominal}
\end{table}

Finally, Table~\ref{tab:real_nominal} gives the unperturbed control condition on the same rig and the same frozen checkpoint. The fine-tuned policy is competent when nothing is injected, so the baseline collapse in Figure~\ref{fig:real_success} is attributable to the perturbation rather than to a weak backbone.

\bibliography{aaai2027}

@article{gu2026safe,
  title={Safe: Multitask failure detection for vision-language-action models},
  author={Gu, Qiao and Ju, Yuanliang and Sun, Shengxiang and Gilitschenski, Igor and Nishimura, Haruki and Itkina, Masha and Shkurti, Florian},
  journal={Advances in Neural Information Processing Systems},
  volume={38},
  pages={40041--40076},
  year={2026}
}

@article{pi0,
  title={$\pi$0: A Vision-Language-Action Flow Model for General Robot Control},
  author={Kevin Black and Noah Brown and Danny Driess and Adnan Esmail and Michael Equi and Chelsea Finn and Niccolo Fusai and Lachy Groom and Karol Hausman and Brian Ichter and Szymon Jakubczak and Tim Jones and Liyiming Ke and Sergey Levine and Adrian Li-Bell and Mohith Mothukuri and Suraj Nair and Karl Pertsch and Lucy Xiaoyang Shi and James Tanner and Quan Vuong and Anna Walling and Haohuan Wang and Ury Zhilinsky},
  journal={ArXiv},
  year={2024},
  volume={abs/2410.24164},
}

@article{Intelligence202505AV,
  title={$\pi$0.5: a Vision-Language-Action Model with Open-World Generalization},
  author={Physical Intelligence and Kevin Black and Noah Brown and James Darpinian and Karan Dhabalia and Danny Driess and Adnan Esmail and Michael Equi and Chelsea Finn and Niccolo Fusai and Manuel Y. Galliker and Dibya Ghosh and Lachy Groom and Karol Hausman and Brian Ichter and Szymon Jakubczak and Tim Jones and Liyiming Ke and Devin LeBlanc and Sergey Levine and Adrian Li-Bell and Mohith Mothukuri and Suraj Nair and Karl Pertsch and Allen Z. Ren and Lucy Xiaoyang Shi and Laura Smith and Jost Tobias Springenberg and Kyle Stachowicz and James Tanner and Quan Vuong and Homer Rich Walke and Anna Walling and Haohuan Wang and Lili Yu and Ury Zhilinsky},
  journal={ArXiv},
  year={2025},
  volume={abs/2504.16054},
}

@inproceedings{openvla,
  author       = {Moo Jin Kim and
                  Karl Pertsch and
                  Siddharth Karamcheti and
                  Ted Xiao and
                  Ashwin Balakrishna and
                  Suraj Nair and
                  Rafael Rafailov and
                  Ethan Paul Foster and
                  Pannag R. Sanketi and
                  Quan Vuong and
                  Thomas Kollar and
                  Benjamin Burchfiel and
                  Russ Tedrake and
                  Dorsa Sadigh and
                  Sergey Levine and
                  Percy Liang and
                  Chelsea Finn},
  editor       = {Pulkit Agrawal and
                  Oliver Kroemer and
                  Wolfram Burgard},
  title        = {OpenVLA: An Open-Source Vision-Language-Action Model},
  booktitle    = {Conference on Robot Learning, 6-9 November 2024, Munich, Germany},
  series       = {Proceedings of Machine Learning Research},
  volume       = {270},
  pages        = {2679--2713},
  publisher    = {{PMLR}},
  year         = {2024},
  url          = {https://proceedings.mlr.press/v270/kim25c.html},
  bibsource    = {dblp computer science bibliography, https://dblp.org}
}

@article{adapter, 
    title={VLA-Adapter: An effective paradigm for tiny-scale vision-language-action model}, volume={40}, ISSN={2159-5399}, url={http://dx.doi.org/10.1609/aaai.v40i22.38931}, DOI={10.1609/aaai.v40i22.38931}, number={22}, journal={Proceedings of the AAAI Conference on Artificial Intelligence}, author={Wang, Yihao and Ding, Pengxiang and Li, Lingxiao and Cui, Can and Ge, Zirui and Tong, Xinyang and Song, Wenxuan and Zhao, Han and Zhao, Wei and Hou, Pengxu and Huang, Siteng and Tang, Yifan and Wang, Wenhui and Zhang, Ru and Liu, Jianyi and Wang, Donglin}, year={2026}, pages={18638–18646} }

@misc{youliang2026NVIDIA,
	author = {NVIDIA},
    title={NVIDIA Isaac GR00T N1.7 - A Foundation Model for Generalist Robots.},
	year = {2026},
	month = {jun 30},
	title = {NVIDIA/{Isaac}-{GR00T}},
	url = {https://github.com/NVIDIA/Isaac-GR00T},
	howpublished = {https://github.com/NVIDIA/Isaac-GR00T},
}

@article{kim2025fine,
  title={Fine-Tuning Vision-Language-Action Models: Optimizing Speed and Success},
  author={Kim, Moo Jin and Finn, Chelsea and Liang, Percy},
  journal={arXiv preprint arXiv:2502.19645},
  year={2025}
}

@inproceedings{robustvla,
title={On Robustness of Vision-Language-Action Model against Multi-Modal Perturbations},
author={Jianing Guo and Zhenhong Wu and Chang Tu and Yiyao Ma and Xiangqi Kong and Zhiqian Liu and Jiaming Ji and Shuning Zhang and Yuanpei Chen and Kai Chen and Qi Dou and Yaodong Yang and Xianglong Liu and Huijie Zhao and Weifeng Lv and Simin Li},
booktitle={The Fourteenth International Conference on Learning Representations},
year={2026},
url={https://openreview.net/forum?id=cS6xizdYD5}
}

@inproceedings{zheng2026xvla,
title={X-{VLA}: Soft-Prompted Transformer as Scalable Cross-Embodiment Vision-Language-Action Model},
author={Jinliang Zheng and Jianxiong Li and Zhihao Wang and Dongxiu Liu and Xirui Kang and Yuchun Feng and Yinan Zheng and Jiayin Zou and Yilun Chen and Jia Zeng and Tai Wang and Ya-Qin Zhang and Jingjing Liu and Xianyuan Zhan},
booktitle={The Fourteenth International Conference on Learning Representations},
year={2026},
url={https://openreview.net/forum?id=kt51kZH4aG}
}

@InProceedings{rt2,
  title = 	 {RT-2: Vision-Language-Action Models Transfer Web Knowledge to Robotic Control},
  author =       {Zitkovich, Brianna and Yu, Tianhe and Xu, Sichun and Xu, Peng and Xiao, Ted and Xia, Fei and Wu, Jialin and Wohlhart, Paul and Welker, Stefan and Wahid, Ayzaan and Vuong, Quan and Vanhoucke, Vincent and Tran, Huong and Soricut, Radu and Singh, Anikait and Singh, Jaspiar and Sermanet, Pierre and Sanketi, Pannag R. and Salazar, Grecia and Ryoo, Michael S. and Reymann, Krista and Rao, Kanishka and Pertsch, Karl and Mordatch, Igor and Michalewski, Henryk and Lu, Yao and Levine, Sergey and Lee, Lisa and Lee, Tsang-Wei Edward and Leal, Isabel and Kuang, Yuheng and Kalashnikov, Dmitry and Julian, Ryan and Joshi, Nikhil J. and Irpan, Alex and Ichter, Brian and Hsu, Jasmine and Herzog, Alexander and Hausman, Karol and Gopalakrishnan, Keerthana and Fu, Chuyuan and Florence, Pete and Finn, Chelsea and Dubey, Kumar Avinava and Driess, Danny and Ding, Tianli and Choromanski, Krzysztof Marcin and Chen, Xi and Chebotar, Yevgen and Carbajal, Justice and Brown, Noah and Brohan, Anthony and Arenas, Montserrat Gonzalez and Han, Kehang},
  booktitle = 	 {Proceedings of The 7th Conference on Robot Learning},
  pages = 	 {2165--2183},
  year = 	 {2023},
  editor = 	 {Tan, Jie and Toussaint, Marc and Darvish, Kourosh},
  volume = 	 {229},
  series = 	 {Proceedings of Machine Learning Research},
  month = 	 {06--09 Nov},
  publisher =    {PMLR},
  url = 	 {https://proceedings.mlr.press/v229/zitkovich23a.html}
}

@article{Ma_2026,
   title={A Survey on Vision--Language--Action Models for Embodied AI},
   volume={37},
   ISSN={2162-2388},
   url={http://dx.doi.org/10.1109/TNNLS.2025.3650584},
   DOI={10.1109/tnnls.2025.3650584},
   number={7},
   journal={IEEE Transactions on Neural Networks and Learning Systems},
   publisher={Institute of Electrical and Electronics Engineers (IEEE)},
   author={Ma, Yueen and Song, Zixing and Zhuang, Yuzheng and Hao, Jianye and King, Irwin},
   year={2026},
   month=July, pages={3031–3051} }

@article{pumacay2024colosseum,
  title     = {THE COLOSSEUM: A Benchmark for Evaluating Generalization for Robotic Manipulation}, 
  author    = {Pumacay, Wilbert and Singh, Ishika and Duan, Jiafei and Krishna, Ranjay and Thomason, Jesse and Fox, Dieter},
  booktitle = {arXiv preprint arXiv:2402.08191},
  year      = {2024},
}

@INPROCEEDINGS{RACER,
  author={Dai, Yinpei and Lee, Jayjun and Fazeli, Nima and Chai, Joyce},
  booktitle={2025 IEEE International Conference on Robotics and Automation (ICRA)}, 
  title={RACER: Rich Language-Guided Failure Recovery Policies for Imitation Learning}, 
  year={2025},
  volume={},
  number={},
  pages={15657-15664},
  doi={10.1109/ICRA55743.2025.11127799}}

@inproceedings{
duan2025aha,
title={{AHA}: A Vision-Language-Model for Detecting and Reasoning Over Failures in Robotic Manipulation},
author={Jiafei Duan and Wilbert Pumacay and Nishanth Kumar and Yi Ru Wang and Shulin Tian and Wentao Yuan and Ranjay Krishna and Dieter Fox and Ajay Mandlekar and Yijie Guo},
booktitle={The Thirteenth International Conference on Learning Representations},
year={2025},
url={https://openreview.net/forum?id=JVkdSi7Ekg}
}

@inproceedings{
xiao2026selfimproving,
title={Self-Improving Vision-Language-Action Models with Data Generation via Residual {RL}},
author={Wenli Xiao and Haotian Lin and Andy Peng and Haoru Xue and Tairan He and Zhengyi Luo and Yuqi Xie and Fengyuan Hu and Linxi Fan and Guanya Shi and Yuke Zhu},
booktitle={The Fourteenth International Conference on Learning Representations},
year={2026},
url={https://openreview.net/forum?id=eUGoqrZ6Ea}
}

@InProceedings{Fei_2026_CVPR,
    author    = {Fei, Senyu and Wang, Siyin and Shi, Junhao and Dai, Zihao and Cai, Jikun and Qian, Pengfang and Ji, Li and He, Xinzhe and Zhang, Shiduo and Fei, Zhaoye and Fu, Jinlan and Gong, Jingjing and Qiu, Xipeng},
    title     = {LIBERO-Plus: A Progressive Robustness Benchmark for Visual-Language-Action Models},
    booktitle = {Proceedings of the IEEE/CVF Conference on Computer Vision and Pattern Recognition (CVPR)},
    month     = {June},
    year      = {2026},
    pages     = {38574-38583}
}

@inproceedings{
liu2025what,
title={What Can {RL} Bring to {VLA} Generalization? An Empirical Study},
author={Jijia Liu and Feng Gao and Bingwen Wei and Xinlei Chen and Qingmin Liao and Yi Wu and Chao Yu and Yu Wang},
booktitle={The Thirty-ninth Annual Conference on Neural Information Processing Systems},
year={2025},
url={https://openreview.net/forum?id=qmBMPInbZC}
}

@InProceedings{HYDRA,
  title = 	 {HYDRA: Hybrid Robot Actions for Imitation Learning},
  author =       {Belkhale, Suneel and Cui, Yuchen and Sadigh, Dorsa},
  booktitle = 	 {Proceedings of The 7th Conference on Robot Learning},
  pages = 	 {2113--2133},
  year = 	 {2023},
  editor = 	 {Tan, Jie and Toussaint, Marc and Darvish, Kourosh},
  volume = 	 {229},
  series = 	 {Proceedings of Machine Learning Research},
  month = 	 {06--09 Nov},
  publisher =    {PMLR},
  url = 	 {https://proceedings.mlr.press/v229/belkhale23a.html}
}

@InProceedings{Hi_Robot,
  title = 	 {Hi Robot: Open-Ended Instruction Following with Hierarchical Vision-Language-Action Models},
  author =       {Shi, Lucy Xiaoyang and Ichter, Brian and Equi, Michael Robert and Ke, Liyiming and Pertsch, Karl and Vuong, Quan and Tanner, James and Walling, Anna and Wang, Haohuan and Fusai, Niccolo and Li-Bell, Adrian and Driess, Danny and Groom, Lachy and Levine, Sergey and Finn, Chelsea},
  booktitle = 	 {Proceedings of the 42nd International Conference on Machine Learning},
  pages = 	 {54919--54933},
  year = 	 {2025},
  editor = 	 {Singh, Aarti and Fazel, Maryam and Hsu, Daniel and Lacoste-Julien, Simon and Berkenkamp, Felix and Maharaj, Tegan and Wagstaff, Kiri and Zhu, Jerry},
  volume = 	 {267},
  series = 	 {Proceedings of Machine Learning Research},
  month = 	 {13--19 Jul},
  publisher =    {PMLR},
  url = 	 {https://proceedings.mlr.press/v267/shi25d.html}
}

@InProceedings{Zhao_2026_CVPR,
    author    = {Zhao, Ganlong and Tang, Zijia and Chen, Xingping and Kuang, Zhanghui and Tian, Ye and Li, Guanbin},
    title     = {FLARE: A Failure-Aware Framework for Autonomous Correction and Recovery in Visual-Language Robotic Manipulation},
    booktitle = {Proceedings of the IEEE/CVF Conference on Computer Vision and Pattern Recognition (CVPR)},
    month     = {June},
    year      = {2026},
    pages     = {22391-22401}
}

@article{liu2024rdt,
  title={RDT-1B: a Diffusion Foundation Model for Bimanual Manipulation},
  author={Liu, Songming and Wu, Lingxuan and Li, Bangguo and Tan, Hengkai and Chen, Huayu and Wang, Zhengyi and Xu, Ke and Su, Hang and Zhu, Jun},
  journal={arXiv preprint arXiv:2410.07864},
  year={2024}
}

@article{lin2025failsafe,
  title={Failsafe: Reasoning and recovery from failures in vision-language-action models},
  author={Lin, Zijun and Duan, Jiafei and Fang, Haoquan and Fox, Dieter and Krishna, Ranjay and Tan, Cheston and Wen, Bihan},
  journal={arXiv preprint arXiv:2510.01642},
  year={2025}
}

@article{li2026learning,
  title={Learning actionable manipulation recovery via counterfactual failure synthesis},
  author={Li, Dayou and Lei, Jiuzhou and Wang, Hao and Liu, Lulin and Yang, Yunhao and Wang, Zihan and Liu, Bangya and Zheng, Minghui and Fan, Zhiwen},
  journal={arXiv preprint arXiv:2603.13528},
  year={2026}
}

@article{ma2026cyclevla,
  title={CycleVLA: Proactive Self-Correcting Vision-Language-Action Models via Subtask Backtracking and Minimum Bayes Risk Decoding},
  author={Ma, Chenyang and Yang, Guangyu and Lu, Kai and Xu, Shitong and Byrne, Bill and Trigoni, Niki and Markham, Andrew},
  journal={arXiv preprint arXiv:2601.02295},
  year={2026}
}

@article{liu2023reflect,
  title={Reflect: Summarizing robot experiences for failure explanation and correction},
  author={Liu, Zeyi and Bahety, Arpit and Song, Shuran},
  journal={arXiv preprint arXiv:2306.15724},
  year={2023}
}

@article{ye2025robofac,
  title={RoboFAC: A Comprehensive Framework for Robotic Failure Analysis and Correction}, 
      author={Zewei Ye and Weifeng Lu and Minghao Ye and Tao Lin and Shuo Yang and Junchi Yan and Bo Zhao},
      year={2026},
      eprint={2505.12224},
      archivePrefix={arXiv},
      primaryClass={cs.RO},
      url={https://arxiv.org/abs/2505.12224}, 
}

@misc{hu2026recovla,
      title={ReCoVLA: VLM-Guided Reward Compilation for Failure Recovery in Vision-Language-Action Policies}, 
      author={Haodi Hu and Chung-Ta Huang and Jing Liu and Ye Wang and Kei Suzuki and Matthew Brand and Toshiaki Koike-Akino},
      year={2026},
      eprint={2606.09630},
      archivePrefix={arXiv},
      primaryClass={cs.RO},
      url={https://arxiv.org/abs/2606.09630}, 
}

@inproceedings{xu2025can,
  title={Can we detect failures without failure data? uncertainty-aware runtime failure detection for imitation learning policies},
  author={Xu, Chen and Nguyen, Tony Khuong and Dixon, Emma and Rodriguez, Christopher and Miller, Patrick and Lee, Robert and Shah, Paarth and Ambrus, Rares Andrei and Nishimura, Haruki and Itkina, Masha},
  booktitle={Robot Evaluation for the Real World},
  year={2025}
}

@article{wu2026flowpro,
  title={FlowPRO: Reward-Free Reinforced Fine-Tuning of Flow-Matching VLAs via Proximalized Preference Optimization},
  author={Wu, Yihao and Zhang, He and Tan, Junbo and Wang, Xueqian and Zhang, Zhengyou},
  journal={arXiv preprint arXiv:2606.05468},
  year={2026}
}

@article{liufu2026repo,
  title={RePO-VLA: Recovery-Driven Policy Optimization for Vision-Language-Action Models},
  author={Liufu, Weijia and Guo, Xiaoyu and Chen, Ruiyi and Liu, Jingzhi and Zhang, Kaidong and Liang, Xiwen and Lin, Jianqi and Sun, Dawei and Wang, Yuze and Xu, Rongtao and others},
  journal={arXiv preprint arXiv:2605.09410},
  year={2026}
}

@misc{berk2026recall,
      title={RECALL: Recovery Experience Collection for Active Lifelong Learning in Vision-Language-Action Models}, 
      author={Ulas Berk Karli and Tesca Fitzgerald},
      year={2026},
      eprint={2606.23617},
      archivePrefix={arXiv},
      primaryClass={cs.RO},
      url={https://arxiv.org/abs/2606.23617}, 
}

@article{zeng2025diagnose,
  title={Diagnose, Correct, and Learn from Manipulation Failures via Visual Symbols},
  author={Zeng, Xianchao and Zhou, Xinyu and Li, Youcheng and Shi, Jiayou and Li, Tianle and Chen, Liangming and Ren, Lei and Li, Yong-Lu},
  journal={arXiv preprint arXiv:2512.02787},
  year={2025}
}

@misc{shin2026back,
      title={Back to the Familiar Future: Failure Recovery for VLA Policies via Pre-Imagined Milestone Selection}, 
      author={Suyeon Shin and Juwon Kim and Hyeonbin Park and Hyunseo Kim and Hyundo Lee and Hyung-Sin Kim and Byoung-Tak Zhang},
      year={2026},
      eprint={2606.09258},
      archivePrefix={arXiv},
      primaryClass={cs.RO},
      url={https://arxiv.org/abs/2606.09258}, 
}

@article{zhao2026retrieve,
  title={Retrieve-then-Steer: Online Success Memory for Test-Time Adaptation of Generative VLAs},
  author={Zhao, Jianchao and Yang, Huoren and Hu, Yusong and Gao, Yuyang and Ou, Qiguan and Wan, Cong and Dong, SongLin and Ma, Zhiheng and Gong, Yihong},
  journal={arXiv preprint arXiv:2605.10094},
  year={2026}
}

@article{liu2023libero,
  title={LIBERO: Benchmarking Knowledge Transfer for Lifelong Robot Learning},
  author={Liu, Bo and Zhu, Yifeng and Gao, Chongkai and Feng, Yihao and Liu, Qiang and Zhu, Yuke and Stone, Peter},
  journal={arXiv preprint arXiv:2306.03310},
  year={2023}
}

@article{tao2024maniskill3,
  title={ManiSkill3: GPU Parallelized Robotics Simulation and Rendering for Generalizable Embodied AI},
  author={Stone Tao and Fanbo Xiang and Arth Shukla and Yuzhe Qin and Xander Hinrichsen and Xiaodi Yuan and Chen Bao and Xinsong Lin and Yulin Liu and Tse-kai Chan and Yuan Gao and Xuanlin Li and Tongzhou Mu and Nan Xiao and Arnav Gurha and Viswesh Nagaswamy Rajesh and Yong Woo Choi and Yen-Ru Chen and Zhiao Huang and Roberto Calandra and Rui Chen and Shan Luo and Hao Su},
  journal = {Robotics: Science and Systems},
  year={2025},
}

@InProceedings{xiang2020sapien,
author = {Xiang, Fanbo and Qin, Yuzhe and Mo, Kaichun and Xia, Yikuan and Zhu, Hao and Liu, Fangchen and Liu, Minghua and Jiang, Hanxiao and Yuan, Yifu and Wang, He and Yi, Li and Chang, Angel X. and Guibas, Leonidas J. and Su, Hao},
title = {{SAPIEN}: A SimulAted Part-based Interactive ENvironment},
booktitle = {The IEEE Conference on Computer Vision and Pattern Recognition (CVPR)},
month = {June},
year = {2020}}

@inproceedings{zhu2020robosuite,
  title={robosuite: A Modular Simulation Framework and Benchmark for Robot Learning},
  author={Yuke Zhu and Josiah Wong and Ajay Mandlekar and Roberto Mart\'{i}n-Mart\'{i}n and Abhishek Joshi and Soroush Nasiriany and Yifeng Zhu and Kevin Lin},
  booktitle={arXiv preprint arXiv:2009.12293},
  year={2020}
}

@inproceedings{todorov2012mujoco,
  title={MuJoCo: A physics engine for model-based control},
  author={Todorov, Emanuel and Erez, Tom and Tassa, Yuval},
  booktitle={2012 IEEE/RSJ International Conference on Intelligent Robots and Systems},
  pages={5026--5033},
  year={2012},
  organization={IEEE},
  doi={10.1109/IROS.2012.6386109}
}


\end{document}